%% file: main.tex
\documentclass[11pt]{article}

\usepackage[final]{acl}

\usepackage{times}
\usepackage{latexsym}

\usepackage[T1]{fontenc}

\usepackage[utf8]{inputenc}

\usepackage{microtype}

\usepackage{inconsolata}

\usepackage{graphicx}

\usepackage{hyperref}
\usepackage{url}
\usepackage{booktabs}       
\usepackage{amsfonts}       
\usepackage{nicefrac}       
\usepackage[table]{xcolor}         
\usepackage{amsmath}
\usepackage{amssymb}
\usepackage{mathtools}
\usepackage{amsthm}
\usepackage{bm}
\usepackage{booktabs}
\usepackage{multirow}
\usepackage[most]{tcolorbox}
\usepackage{enumitem}
\usepackage{subcaption}
\usepackage{rotating}
\usepackage[normalem]{ulem}
\usepackage{pifont}
\useunder{\uline}{\ul}{}
\usepackage{bbm}
\usepackage{algorithm}
\usepackage{algorithmic}

\usepackage{balance}
\usepackage{sidecap}

\newcommand{\AlgStep}[1]{%
  \newline \colorbox{gray!20}{\parbox{\dimexpr\linewidth-2\fboxsep}{#1}}%
  \vspace{-7pt}
}

\title{Subspace Alignment for Vision-Language Model Test-time Adaptation}

\author{
 \textbf{Zhichen Zeng\textsuperscript{1}},
 \textbf{Wenxuan Bao\textsuperscript{1}},
 \textbf{Xiao Lin\textsuperscript{1}},
 \textbf{Ruizhong Qiu\textsuperscript{1}},
 \textbf{Tianxin Wei\textsuperscript{1}},
 \textbf{Xuying Ning\textsuperscript{1}},
\\
 \textbf{Yuchen Yan\textsuperscript{2}},
 \textbf{Chen Luo\textsuperscript{2}},
 \textbf{Monica Xiao Cheng\textsuperscript{2}},
 \textbf{Jingrui He\textsuperscript{1}},
 \textbf{Hanghang Tong\textsuperscript{1}}
\\
\\
 \textsuperscript{1}University of Illinois Urbana-Champaign,
 \textsuperscript{2}Amazon
\\
 \small{
   \textbf{Correspondence:} \href{mailto:htong@illinois.edu}{htong@illinois.edu}
 }
}

\input{sections/hd}

\begin{document}
\maketitle

\input{sections/0-abs}
\input{sections/1-intro}

\input{sections/2-related}
\input{sections/3-method}

\input{sections/4-exp}

\input{sections/5-con}

\newpage
\bibliography{main}

\input{sections/app}

\end{document}

%% file: sections/hd.tex
\newcommand{\beqa}{\begin{eqnarray}}
\newcommand{\eeqa}{\end{eqnarray}}
\newcommand{\beq}{\begin{equation}}
\newcommand{\eeq}{\end{equation}}
\newcommand{\ben}{\begin{enumerate}}
\newcommand{\een}{\end{enumerate}}
\newcommand{\bit}{\begin{itemize}}
\newcommand{\eit}{\end{itemize}}
\newcommand{\bi}{\begin{itemize} \item}
\newcommand{\ei}{\end{itemize}}

\newcommand{\begindef}{\begin{Definition} \rm}
\newcommand{\beginexa}{\begin{Example} \rm}
\newcommand{\beginthe}{\begin{Theorem} \rm}
\newcommand{\beginpro}{\begin{Proposition} \rm}
\newcommand{\beginlem}{\begin{Lemma} \rm}
\newcommand{\begincon}{\begin{Conjecture} \rm}
\newcommand{\begincor}{\begin{Corollary} \rm}

\newcommand{\bluecell}[1]{\cellcolor[HTML]{CFE2F3}#1}
\newcommand{\yellowcell}[1]{\cellcolor[HTML]{FFF2CC}#1}
\newcommand{\redcell}[1]{\cellcolor[HTML]{FADBD8}#1}

\newcommand{\eat}[1]{}

\def\argmax{\mathop{\arg\max}}

\def\V{\mathcal{V}}
\def\T{\mathcal{T}}

\DeclareMathOperator{\Tr}{Tr}

\def\algname{\texttt{SubTTA}}

%% file: sections/0-abs.tex
\begin{abstract}
    Vision-language models (VLMs), despite their extraordinary zero-shot capabilities, are vulnerable to distribution shifts.
    Test-time adaptation (TTA) emerges as a predominant strategy to adapt VLMs to unlabeled test data on the fly.
    However, existing TTA methods heavily rely on zero-shot predictions as pseudo-labels for self-training, which can be unreliable under distribution shifts and misguide adaptation due to two fundamental limitations.
    First (\textit{Modality Gap}), distribution shifts induce gaps between visual and textual modalities, making cross-modal relations inaccurate.
    Second (\textit{Visual Nuisance}), visual embeddings encode rich but task-irrelevant noise that often overwhelms task-specific semantics under distribution shifts.
    To address these limitations, we propose \algname, which aligns the semantic subspaces of both modalities to enhance zero-shot predictions to better guide the TTA process.
    To bridge the modality gap, \algname\ extracts the principal subspaces of both modalities and aligns the visual manifold to the textual semantic anchor by minimizing their chordal distance.
    To eliminate visual nuisance, \algname\ projects the aligned visual features onto the task-specific textual subspace, which filters out task-irrelevant noise by constraining visual embeddings within the valid semantic span, and standard TTA is further performed on the purified space to refine the decision boundaries.
    Extensive experiments on various benchmarks and VLM architectures demonstrate the effectiveness of \algname, yielding an average improvement of 2.24\% over state-of-the-art TTA methods.
    Our code is available at \url{https://github.com/zhichenz98/SubTTA_EMNLP26}.
\end{abstract}

%% file: sections/1-intro.tex
\section{Introduction}

Pretrained Vision-Language Models (VLMs), such as CLIP~\cite{radford2021learning} and ALIGN~\cite{jia2021scaling}, have demonstrated extraordinary zero-shot capabilities across diverse downstream tasks, ranging from image classification~\cite{radford2021learning,addepalli2024leveraging} to image captioning~\cite{chen2022pali,yu2022coca} and visual question-answering~\cite{yu2023self,huynh2025visual,lin2025moralise}. 
The success stems from the expressive joint embedding space, where visual representations are globally aligned with rich linguistic concepts, enabling task descriptions in natural language to directly retrieve task-relevant visual semantics.

Despite these capabilities, VLMs often struggle when deployed in open-world scenarios which are characterized by distribution shifts, such as image corruptions~\cite{hendrycks2019benchmarking} or stylistic changes~\cite{patashnik2021styleclip}.
These shifts distort the vision-language embedding space and degrade the reliability of zero-shot predictions.
To mitigate this, test-time adaptation (TTA) has emerged as a predominant paradigm to adapt pre-trained VLMs to unlabeled test data on the fly~\cite{osowiechi2024watt,maharana2025batclip,bao2025mint}. 
Notably, most existing TTA approaches, either training-free methods utilizing memory banks~\cite{zhang2024dual,li2025efficient} or training-based methods optimizing pseudo labels~\cite{zhang2022memo,shu2022test}, heavily rely on the VLMs' raw zero-shot predictions to guide the adaptation process, which can be precarious when the aligned space is disrupted.
We attribute the failure of standard TTA to two fundamental limitations under distribution shifts, namely \textit{modality gap} and \textit{visual nuisance}.

\begin{figure}[t]
    \centering
    \begin{subfigure}{.41\linewidth}
        \centering
        \includegraphics[width=\textwidth]{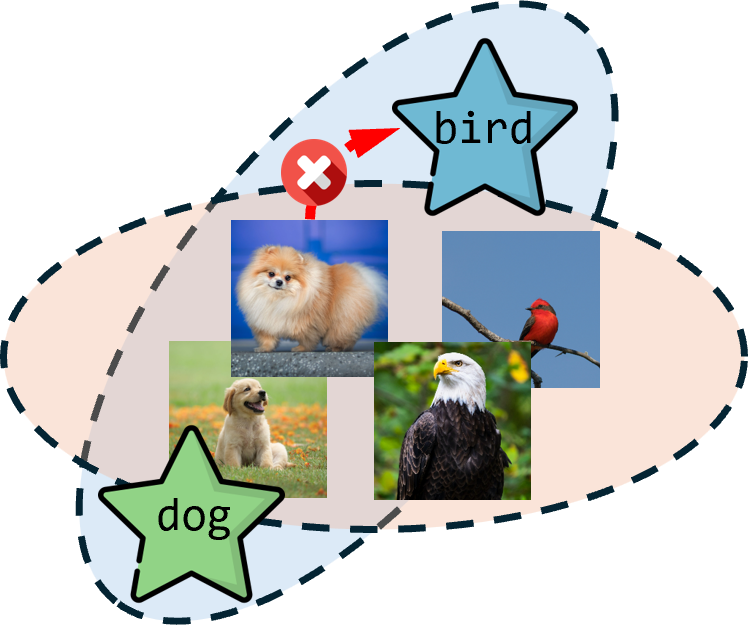}
        \caption{Modality Gap}
        \label{fig:teaser_gap}
    \end{subfigure}
    \begin{subfigure}{.57\linewidth}
        \centering
        \includegraphics[width=\textwidth]{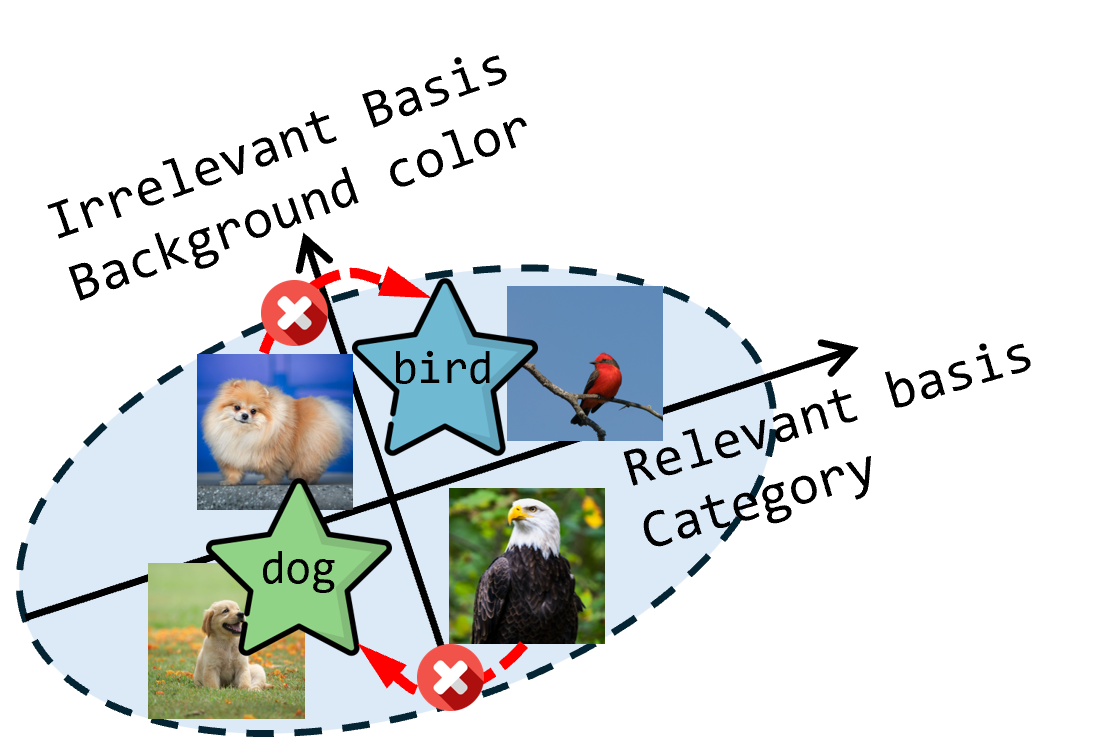}
        \caption{Visual Nuisance}
        \label{fig:teaser_nuisance}
    \end{subfigure}
    \caption{Failure modes of zero-shot prediction. (a) Modality gap: Visual features drift away from the textual manifold. A \texttt{dog} image shifts closer to the \texttt{bird} anchor. (b) Visual Nuisance: Task-irrelevant noise  overshadows core semantics. a \texttt{dog} is misclassified as \texttt{bird} due to spurious correlation with the blue sky.}
    \label{fig:teaser}
\end{figure}
\begin{figure}[t]
    \centering
    \includegraphics[width=\linewidth]{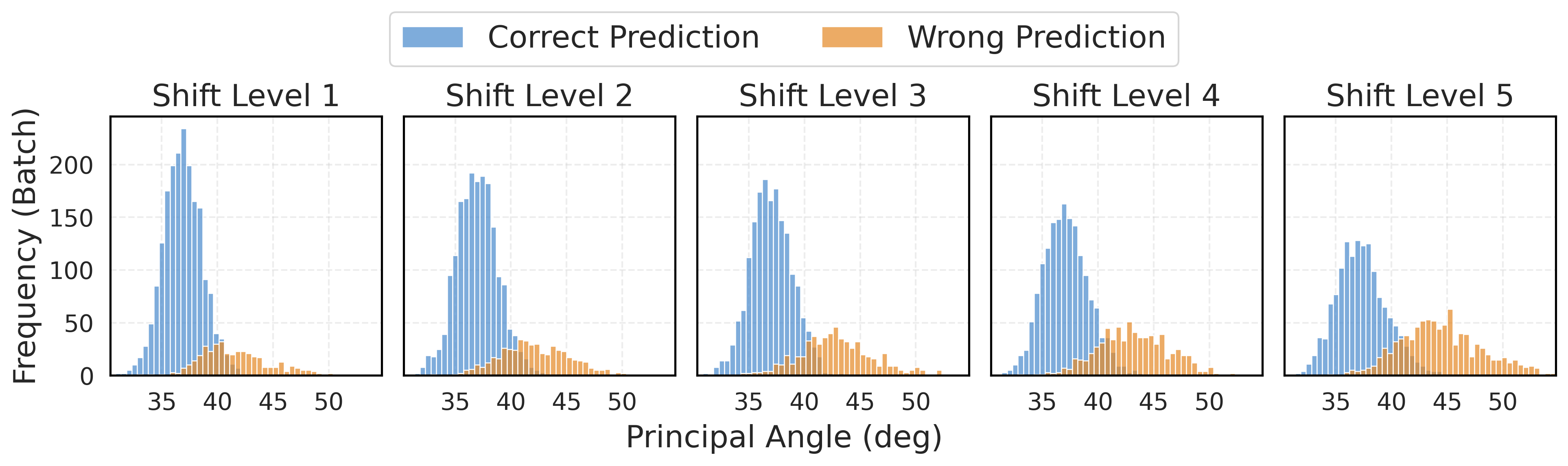}
    \caption{Principal angles ($\downarrow$). \textcolor{blue}{Correct predictions} exhibit consistently smaller principal angles than \textcolor{orange}{mispredictions}. Increased shift level results in larger angles and more mispredictions.}
    \label{fig:angle}
\end{figure}

First (\textit{Modality Gap}), distribution shifts induce a global drift of the visual manifold relative to the textual manifold.
Intuitively, as shown in Figure~\ref{fig:teaser_gap}, the modality gap causes visual features (e.g., \texttt{dog}) to drift toward incorrect textual anchors (e.g., \texttt{bird}), leading to incorrect zero-shot predictions.
To validate this empirically, we analyze the visual-textual principal angles under different shift levels in Figure~\ref{fig:angle}.
We observe that correct predictions consistently exhibit smaller principal angles than mispredictions, and that, as the shift level increases to the right side, the entire distribution shifts toward larger angles accompanied by a surge in errors. 
This geometric divergence indicates that the pre-trained vision-language alignment is structurally broken, creating a modality gap where the visual feature space is globally rotated away from the textual anchor space.

\begin{figure}[t]
    \centering
    \includegraphics[width=\linewidth]{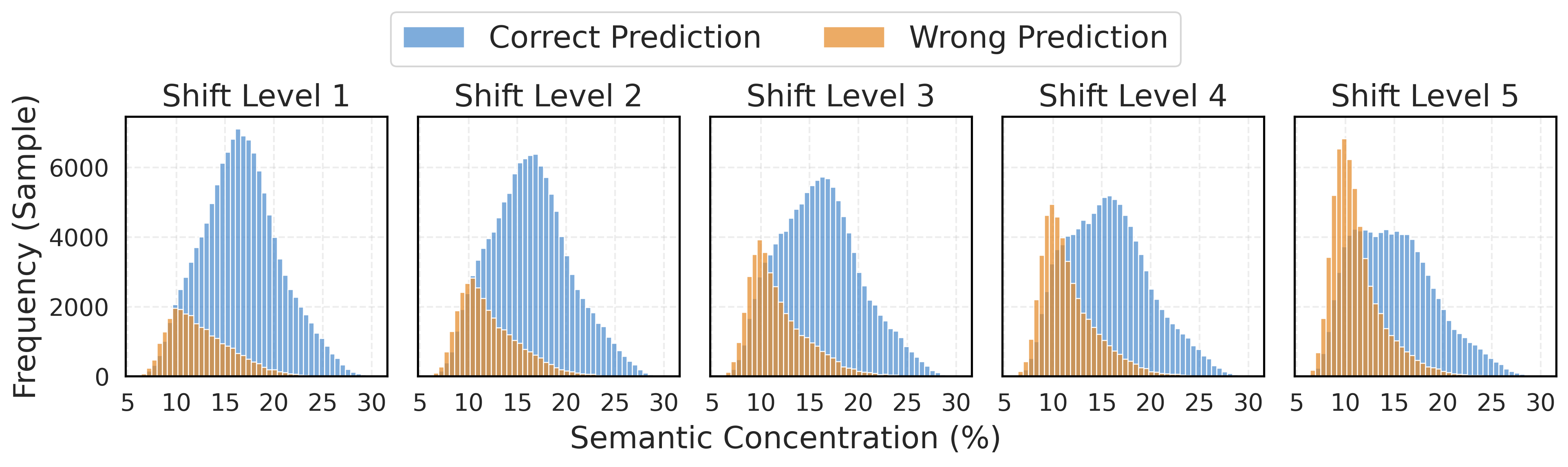}
    \caption{Semantic concentration ($\uparrow$). \textcolor{blue}{Correct predictions} exhibit markedly higher semantic concentration than \textcolor{orange}{mispredictions}. Increased shift level results in lower concentration and more mispredictions.}
    \label{fig:semantic}
\end{figure}

Second (\textit{Visual Nuisance}), unlike compact textual anchors, visual embeddings encode rich but task-irrelevant information.
As illustrated in Figure~\ref{fig:teaser_nuisance}, irrelevant nuisances often overshadow core semantics: for example, a \texttt{dog} on a blue background may be misclassified as a \texttt{bird} due to the spurious correlation between the sky color and the \texttt{bird} class.
We quantify this phenomenon via semantic concentration, measured by the ratio of visual energy projected onto the textual subspace relative to the raw embedding, in Figure~\ref{fig:semantic}.
Correct samples exhibit markedly higher semantic concentration than mispredictions, while increasing shift levels push the distributions toward lower concentration, revealing that task-relevant components are increasingly overshadowed by nuisance dimensions.
Consequently, pseudo-labels derived from raw visual embeddings are heavily contaminated by irrelevant noise (e.g., background clutter or domain-specific styles), causing TTA to reinforce prediction errors rather than correct them.

To address these challenges, we propose \algname, a novel TTA framework grounded in subspace alignment.
First (\textit{Geometric Alignment}), to bridge the modality gap, we rectify the global drift of the visual manifold. We construct compact principal subspaces for both modalities via eigendecomposition.
Treating the textual basis as an anchor, we geometrically align the visual subspace to it by minimizing the chordal distance. 
This step recalibrates the visual feature space to match the pre-trained vision-language geometry.
Second (\textit{Semantic Projection}), to eliminate visual nuisance, we project the aligned visual embeddings onto the task-specific textual subspace.
The projection acts as a semantic filter which constrains visual features to lie within the semantic span defined by the textual basis, effectively discarding irrelevant noise and recovering the submerged semantic signal.
Finally, standard self-training objectives are applied within this purified space to sharpen decision boundaries.

Our contributions are summarized as follows: 
\begin{itemize}[noitemsep, topsep=0pt]
    \item \textbf{Analysis.} We provide a novel perspective on VLM TTA failure, identifying two barriers, namely modality gap and visual nuisance.
    \item \textbf{Method.} We propose \algname, a subspace-centric TTA framework to align visual-textual subspaces and filter visual nuisance via semantic projection, ensuring pseudo label quality.
    \item \textbf{Evaluation.} Extensive experiments on diverse benchmarks and VLM architectures demonstrate that \algname\ significantly outperforms state-of-the-art TTA methods.
\end{itemize}

%% file: sections/2-related.tex
\section{Related Works}

\paragraph{Vision-Language Model Test-time Adaptation.}
Existing VLM TTA methods are broadly categorized into training-based and training-free approaches. \emph{Training-based} methods update parameters or prompts via self-supervised objectives, such as entropy minimization~\cite{wang2020tent,zhang2022memo}, robust loss functions~\cite{rusak2022if,yuan2023robust}, or context prompt tuning~\cite{shu2022test}. Recent variants further explore cluster-based refinement~\cite{maharana2025batclip,bao2025mint}, model merging~\cite{osowiechi2024watt}, and negative augmentation~\cite{deng2025panda}. Conversely, \emph{training-free} methods avoid gradient updates for better efficiency, calibrating distributions via logit/prototype adjustments~\cite{zhang2024dual,li2025efficient,karmanov2024efficient,huang2026prototype} or closed-form feature alignment~\cite{farina2024frustratingly,dobler2024lost}. Despite their effectiveness, both paradigms typically rely on raw zero-shot predictions as pseudo-labels. These can be highly noisy under distribution shifts, risking catastrophic failures. \algname\ addresses this by intrinsically refining zero-shot predictions to enable reliable adaptation.

\paragraph{Geometric Adaptation.}
Exploiting feature space geometry is a growing trend for bridging modality gaps. However, prior subspace methods typically focus on few-shot~\cite{zhu2024selective,simon2020adaptive,wang2025complementary} or uni-modal~\cite{adachitest,fernando2014subspace} settings, often utilizing rigid rank-to-rank metrics~\cite{fernando2014subspace,sun2016deep}. For cross-modal VLM adaptation, STS~\cite{dafnis2026test} aligns textual embeddings to visual features, which inadvertently risks accommodating task-irrelevant visual nuisances. 
In contrast, \algname\ differentiates itself in three key aspects.
First (\textit{Task}), \algname\ targets the challenging zero-shot cross-modal classification; Second (\textit{Metric}), by optimizing the basis-invariant chordal distance rather than rigid rank-to-rank metrics, we avoid false correspondences caused by nuisance factors; Third (\textit{Approach}), we explicitly filter visual nuisances via semantic projection, offering a plug-and-play module that integrates seamlessly with existing TTA frameworks.

%% file: sections/3-method.tex
\section{Methodology}\label{sec:method}

In this section, we introduce our proposed \algname\ to improve TTA performance via subspace alignment.
We first introduce preliminaries in Section~\ref{sec:method-pre}.
Afterwards, we introduce two key components, namely geometric alignment and semantic projection, in Sections~\ref{sec:method-align} and \ref{sec:method-proj}, respectively.
\begin{figure*}[t]
    \centering
    \includegraphics[width=\textwidth]{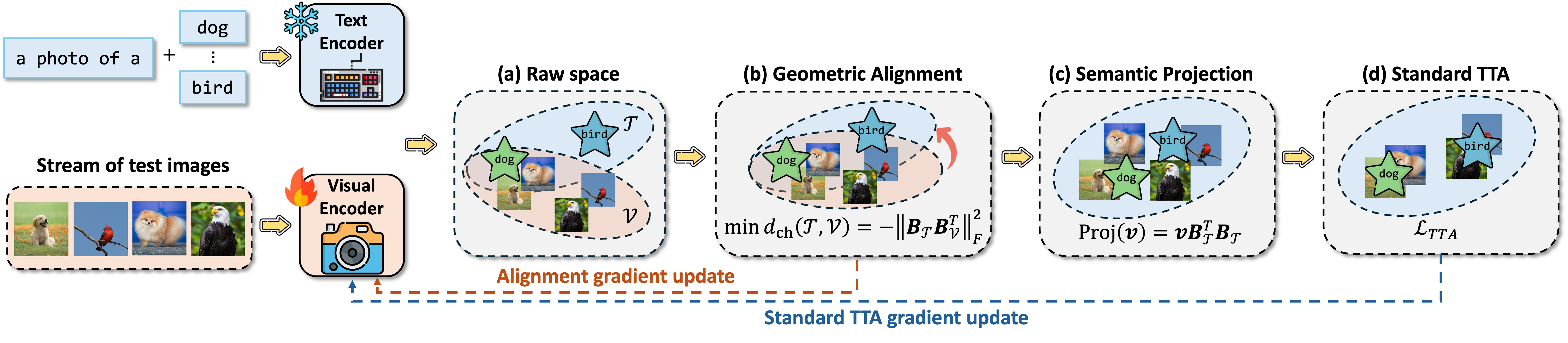}
    \vspace{-20pt}
    \caption{Overview of \algname. (a) VLM encodes images and textual prompts into a shared raw space. (b) Geometric alignment alleviates modality gap by minimizing the chordal distance. (c) Semantic projection retains task-relevant information, e.g., category, while filtering out irrelevant ones, e.g., background color. (d) Standard TTA is further performed on the aligned subspace.}
    \vspace{-10pt}
    \label{fig:overview}
\end{figure*}

\subsection{Preliminaries}\label{sec:method-pre}
We denote text space by $\T$ and image space by $\V$.
A VLM consisting of an image encoder $f_v:\V\to\mathbb{R}^d$ and a text encoder $f_t:\T\to\mathbb{R}^d$, which aligns image and text embeddings in a shared space.
For an image classification task with $C$ classes, the text encoder $f_t$ embeds the textual descriptions, e.g., "\texttt{A photo of a <class>}", into text embeddings $\mathbf{t}_1,...,\mathbf{t}_C\in\mathbb{R}^d$.
Given a test image, the image encoder $f_v$ embeds it into an image embedding $\mathbf{v}\in\mathbb{R}^d$, and the prediction corresponds to the class with the highest similarity score, i.e., $\argmax_{c} \mathbf{v}^\top\mathbf{t}_c$.

\subsection{Geometric Alignment}\label{sec:method-align}
High-dimensional VLM embeddings are susceptible to the modality gap under distribution shifts, where structural misalignment leads to unreliable zero-shot predictions.
To mitigate this, we propose to align the visual and textual representations within a refined low-dimensional subspace.

We begin by identifying the principal directions that capture textual and visual semantics via eigendecomposition on feature covariance matrices.
For the textual modality, textual prompts encapsulate rich task-specific semantic information (e.g., class categories), hence their covariance matrix $\mathbf{\Sigma}_\T$ effectively defines the target semantic space.
Formally, given the normalized text embeddings $\mathbf{T} = [\mathbf{t}_1, \dots, \mathbf{t}_C]^\top \in \mathbb{R}^{C \times d}$, the text covariance matrix is computed as $\mathbf{\Sigma}_\T = \mathbf{T}^\top \mathbf{T}\in\mathbb{R}^{d\times d}$.
For the visual modality, test images typically arrive in small batches, which provide insufficient statistics to accurately estimate the global visual distribution, leading to high estimation variance and noise~\cite{bao2025mint}.
To mitigate such instability, we employ an Exponential Moving Average (EMA) to maintain a robust estimate of the visual covariance.
Specifically, we initialize the visual covariance using the textual covariance as a semantic prior, i.e., $\mathbf{\Sigma}_\V\gets\mathbf{\Sigma}_\T$.
For each step $k$, with the normalized image batch $\mathbf{V}^{(k)}=[\mathbf{v}_1, \dots, \mathbf{v}_B]^\top \in \mathbb{R}^{B \times d}$, the visual covariance matrix is gradually updated by
\begin{equation}\label{eq:ema}
    \mathbf{\Sigma}_\V \gets (1 - \alpha) \mathbf{\Sigma}_\V + \alpha (\mathbf{V}^{(k)^\top}\mathbf{V}^{(k)}),
\end{equation}
where $\alpha$ is the momentum coefficient.

To extract the dominant components from the noisy covariance matrices, we perform eigendecomposition to obtain the rank-$r$ approximations
\begin{equation}\label{eq:basis}
    \mathbf{\Sigma}_\T\approx\mathbf{B}_\T^\top \mathbf{\Lambda}_\T\mathbf{B}_\T, \mathbf{\Sigma}_\V\approx\mathbf{B}^{\top}_\V \mathbf{\Lambda}_\V\mathbf{B}_\V,
\end{equation}
where $\mathbf{B}_\T, \mathbf{B}_\V \in \mathbb{R}^{r \times d}$ are the top-$r$ orthonormal eigenvectors corresponding to the principal basis, and $\mathbf{\Lambda}_\T, \mathbf{\Lambda}_\V$ are diagonal matrices containing the corresponding top-$r$ eigenvalues.

After obtaining the principal bases, we follow prior works~\cite{osowiechi2024watt,hakim2025clipartt,bao2025mint} and adapt the normalization layers of the image encoder to align the visual span $\mathbf{B}_\V$ to the textual anchor $\mathbf{B}_\T$.
Conventional metrics, such as Frobenius norm $\|\mathbf{B}_\T-\mathbf{B}_\V\|_F$ and cosine similarity $\sum_{i=1}^r \mathbf{B}_\T(i)^\top\mathbf{B}_\V(i)$, enforce a \emph{rigid rank-to-rank alignment}, i.e., forcing $\mathbf{B}_\T(i)$ to align with $\mathbf{B}_\V(i)$.
However, this can be problematic because the bases in $\mathbf{B}_\T$ and $\mathbf{B}_\V$ are sorted by statistical variance rather than semantic relevance.
For instance, the top-ranked visual basis may capture dominant but task-irrelevant signals (e.g., background intensity or style) rather than target objects.
Consequently, enforcing a rigid rank-to-rank correspondence may erroneously align these nuisances with the primary textual anchors.
Therefore, the distance metric is expected to be solely defined by the subspace geometry while remaining invariant to the specific choice of basis vectors.

To satisfy this property, we employ the chordal distance on the Grassmannian manifold as follows
\begin{equation}
    d_{\text{ch}}(\mathbf{B}_\T, \mathbf{B}_\V) = \sqrt{\sum_{i=1}^r \sin^2\theta_i},
\end{equation}
where $\theta_i$ represents the $i$-th principal angle between subspaces $\operatorname{span}(\mathbf{B}_\T)$ and $\operatorname{span}(\mathbf{B}_\V)$.
Geometrically, minimizing the chordal distance forces the principal angles toward zero, thereby maximizing the overlap between the visual and textual subspaces.
Based on this, we formulate our alignment objective as the squared chordal distance, which can be efficiently computed via the Frobenius norm as follows~\cite{conway1996packing}
\begin{equation}\label{eq:loss-align}
    \mathcal{L}_{\text{align}} =d_{\text{ch}}^2(\mathbf{B}_\T,\mathbf{B}_\V) = r-\|\mathbf{B}_\T \mathbf{B}_\V^\top\|_F^2.
\end{equation}

Note that the chordal distance is defined solely by the principal angles $\{\theta_i\}_{i=1}^r$ between the two subspaces, remaining invariant to the basis choices.
Formally, consider an orthogonal rotation matrix $\mathbf{Q}\in\mathbb{R}^{r\times r}$ with $\mathbf{Q}^\top\mathbf{Q}=\mathbf{I}$ that rotates the visual bases by $\mathbf{Q}\mathbf{B}_\V$, we have
\begin{equation*}
    \begin{aligned}
        \|\mathbf{B}_\T (\mathbf{Q}\mathbf{B}_\V)^\top\|_F^2 &=\Tr\left(\mathbf{B}_\T\mathbf{B}_\V^\top\mathbf{Q}^\top\mathbf{Q}\mathbf{B}_\V\mathbf{B}_\T^\top\right)\\
        &=\Tr\left(\mathbf{B}_\T\mathbf{B}_\V^\top\mathbf{B}_\V\mathbf{B}_\T^\top\right)\\
        &=\|\mathbf{B}_\T \mathbf{B}_\V^\top\|_F^2.
    \end{aligned}
\end{equation*}
This geometric flexibility avoids rigid rank-to-rank alignment, and enables a global matching that allows relevant visual signals to align with the correct textual anchors regardless of their rank order.

\subsection{Semantic Projection}\label{sec:method-proj}
While geometric alignment effectively rectifies the global distribution drift, it preserves the intrinsic structure of the visual features that include the task-irrelevant nuisance.
To eliminate such nuisance, we leverage the textual subspace, which is compact and rich in task-specific semantics, as a hard semantic filter.
Specifically, for image embedding $\mathbf{v}$, the semantic projection is defined as follows

\begin{equation}\label{eq:proj}
\text{Proj}(\mathbf{v}) = \mathbf{v} \mathbf{B}_\T^\top \mathbf{B}_\T.
\end{equation}
This operation effectively discards noises orthogonal to the semantic span, yielding a purified embedding space that facilitates high-quality zero-shot predictions.
Equipped with the purified embeddings, \algname\ can be seamlessly integrated with various TTA objectives (e.g., entropy-based or cluster-based) to refine decision boundaries.

%% file: sections/4-exp.tex
\input{sections/tab_benchmark_16}
\input{sections/tab_imagenet}

\section{Experiments}\label{sec:exp}

\subsection{Experiment Setup}\label{sec:exp-setup}

\paragraph{Datasets and Metric.} We evaluate \algname\ on three corrupted image classification benchmarks: CIFAR-10-C, CIFAR-100-C, and ImageNet-C~\cite{hendrycks2019benchmarking}, each comprising 15 types of visual corruptions. 
To assess robustness against broader shifts, such as adversarial examples and domain gaps, we additionally evaluate on ImageNet-A/R/K~\cite{hendrycks2021many,hendrycks2021natural,wang2019learning}. 
We adopt the classification accuracy as the evaluation metric.

\noindent\textbf{CLIP Models.}
We consider the following widely used CLIP~\cite{radford2021learning} models, including \texttt{ViT-B-16}, \texttt{ViT-B-32}, and \texttt{ViT-L-14}.

\paragraph{Baselines.}
We benchmark \algname\ against state-of-the-art TTA approaches.
\textit{Training-free methods} include memory-based methods (TDA~\cite{karmanov2024efficient}, DMN~\cite{zhang2024dual}, ECALP~\cite{li2025efficient}) that leverage sample similarity to adjust predictions, and augmentation-based methods (VTE~\cite{dobler2024lost}, ZERO~\cite{farina2024frustratingly}) that aggregate image embeddings from multiple augmentations.
\textit{Training-based methods} include entropy-based methods (MEMO~\cite{zhang2022memo}, WATT~\cite{osowiechi2024watt}, RoTTA~\cite{yuan2023robust}, TPT~\cite{shu2022test}, STS~\cite{dafnis2026test}), and cluster-based methods (MINT~\cite{bao2025mint}, BATCLIP~\cite{maharana2025batclip}).

\paragraph{Pipeline}
Following the established TTA protocol~\cite{wang2020tent}, we conduct experiments under the highest severity level (Level 5) to simulate severe shifts.
We adopt an online adaptation setting where the model adapts to a continuous stream of unlabeled test data for each corruption type independently, resetting the model state between corruptions.
For each batch, we perform one gradient update for the alignment loss (Eq.~\eqref{eq:loss-align}) and the TTA loss, respectively.
All experiments are conducted on NVIDIA A100 80GB GPUs.

\subsection{Benchmark Results}\label{sec:exp-benchmark}

We conduct experiments on ViT-B-16 with results in Tables~\ref{tab:benchmark-16} and \ref{tab:imagenet}.
Additional results on ViT-B-32 and ViT-L-14 are reported in Appendix~\ref{app:exp}.

\noindent\textbf{(1) \algname\ establishes a new state-of-the-art with consistent robustness.} 
\algname\ consistently delivers the strongest overall performance across diverse benchmarks.
In fine-grained settings, \algname\ consistently ranks within the Top-3 and achieves the best performance in the majority of scenarios
On the large-scale ImageNet-C dataset, \algname\ achieves a mean accuracy of 32.15\%, surpassing the previous best method \textsc{BATCLIP} (30.06\%) by a significant margin of 2.09\%.
Such superiority is equally pronounced on smaller-resolution benchmarks: \algname\ outperforms the best competitor by 2.54\% on CIFAR-10-C, and 2.08\% on CIFAR-100-C.
Beyond synthetic corruptions, \algname\ consistently enhances base TTA performance against natural distribution shifts and stylistic variations, achieving an average improvement of 0.56\%.
This validates that \algname\ constructs a geometrically rectified feature space that is inherently more robust to varying forms of distribution shifts.

\noindent\textbf{(2) \algname\ avoids negative adaptation and ensures stability.}
Existing baseline methods often suffer from catastrophic failure under large shifts, leading to negative adaptation where performance drops below the source CLIP baseline. 
For instance, we observe that \textsc{TENT} fails to adapt to noise corruptions due to unstable entropy minimization, while \textsc{DMN} often struggles with digital corruptions.
\algname\ exhibits remarkable resilience, consistently outperforming the source CLIP model across every corruption category. 
This stability confirms that our subspace alignment effectively filters out the nuisance factors that typically destabilize other adaptation algorithms.

\subsection{Performance with Various TTA Objectives}\label{sec:exp-comp}

\begin{figure}[t]
    \centering
    \includegraphics[width=\linewidth]{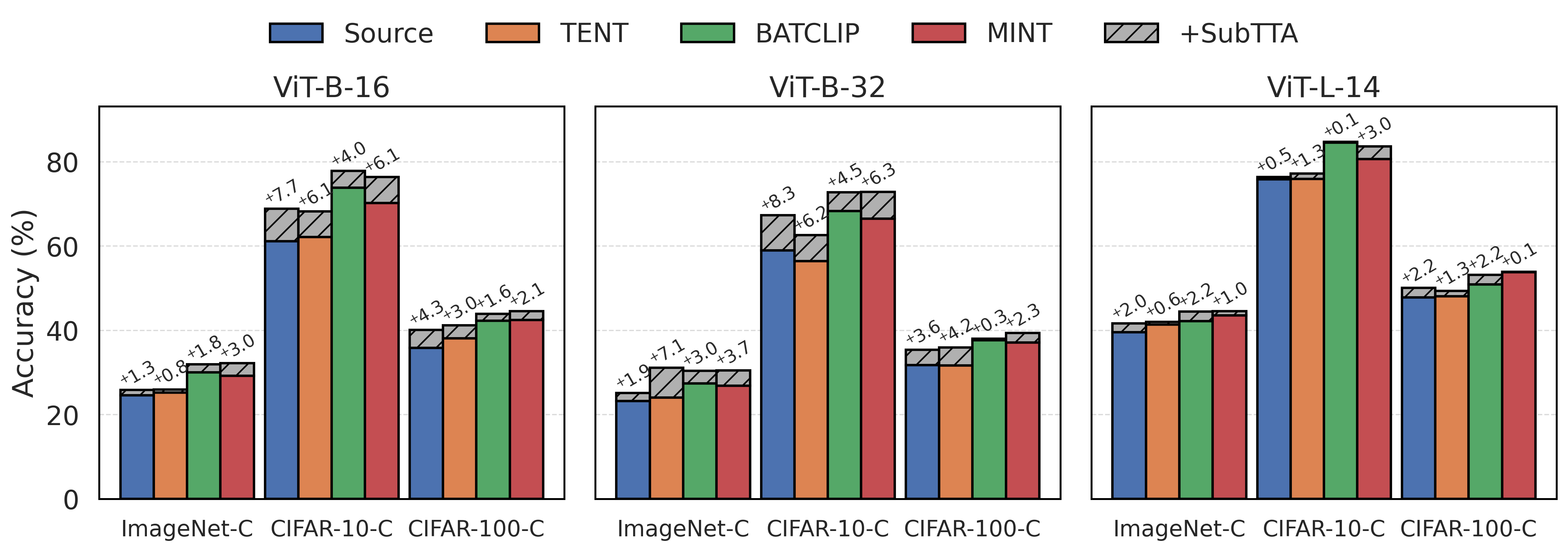}
    \caption{TTA performance w/ and w/o \algname.}
    \label{fig:exp-comp}
\end{figure}

Our proposed \algname\ is compatible with various TTA objectives.
To validate this, we integrate \algname\ with three categories of baseline methods, including Source CLIP model, entropy minimization (TENT), cluster-based methods (MINT, BATCLIP), and the results are shown in Figure~\ref{fig:exp-comp}.

\begin{figure*}[t]
    \centering
    \begin{subfigure}{.32\linewidth}
        \centering
        \includegraphics[width=\textwidth, trim=0 10 0 0, clip]{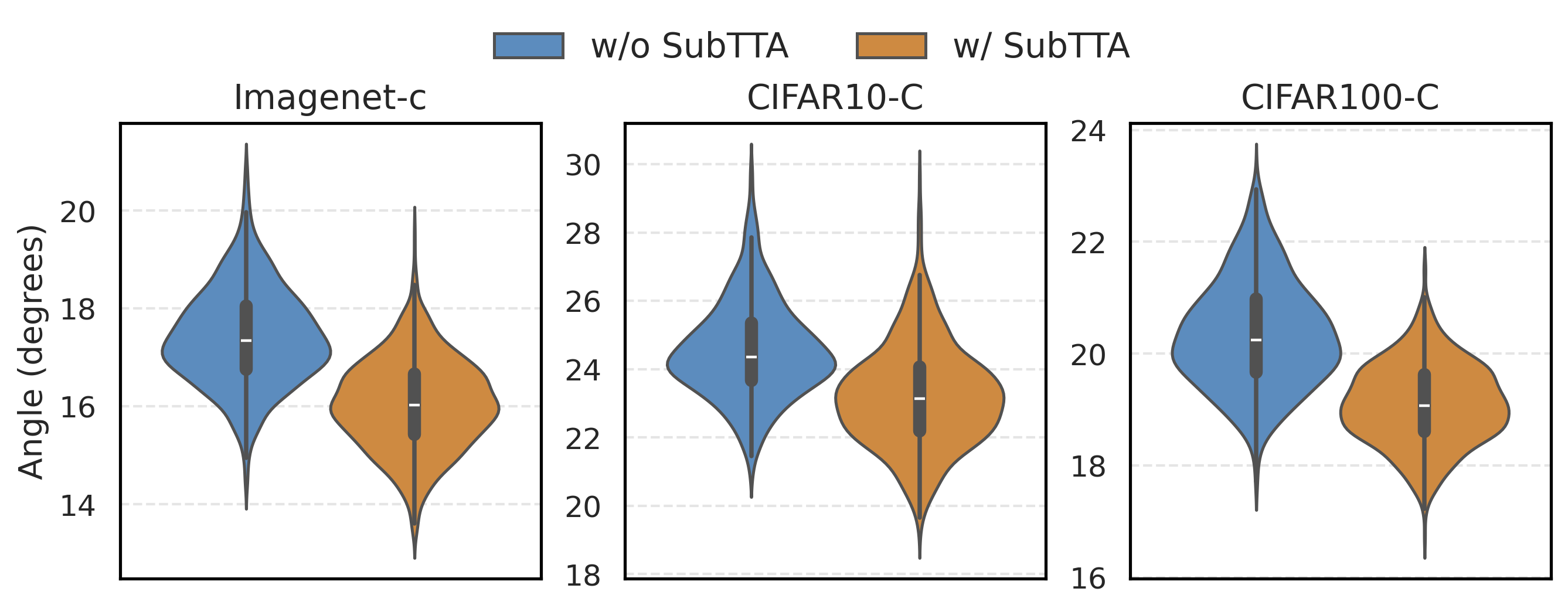}
        \caption{Principal angle.}
        \label{fig:exp-angle}
    \end{subfigure}
    \hfill
    \begin{subfigure}{.32\linewidth}
        \centering
        \includegraphics[width=\textwidth]{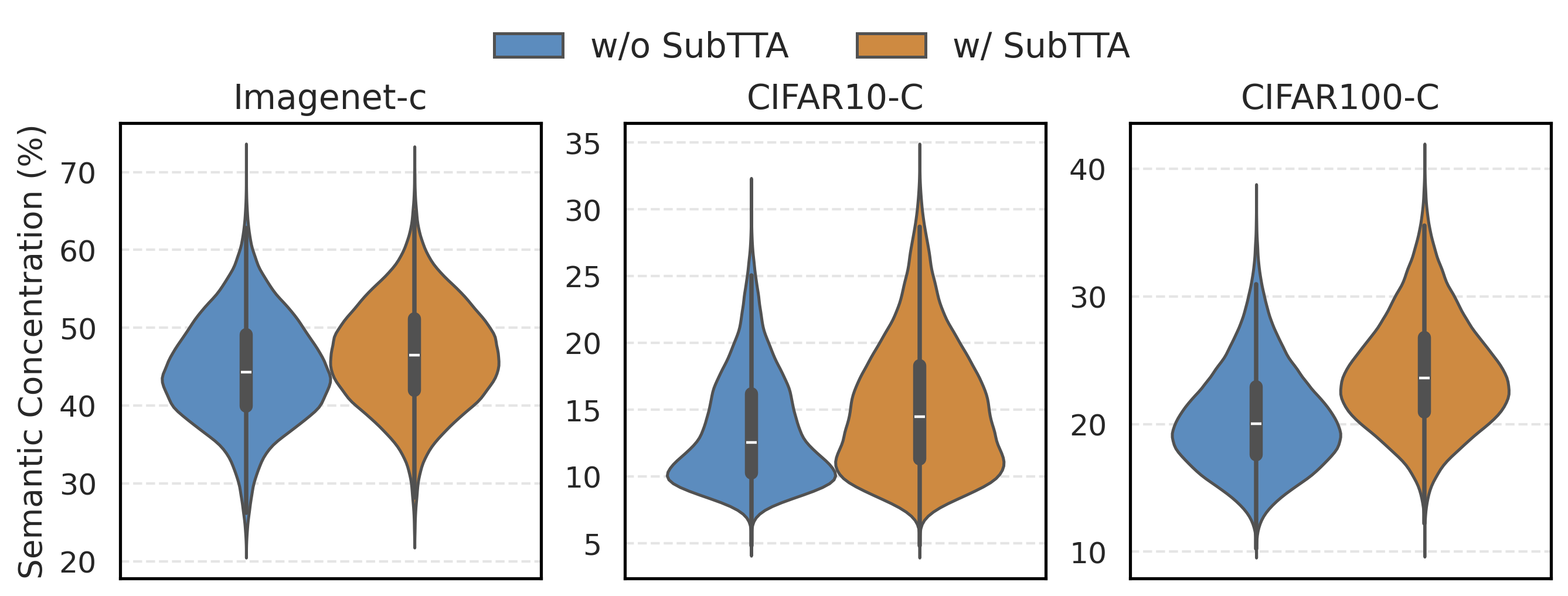}
        \caption{Semantic concentration.}
        \label{fig:exp-nuisance}
    \end{subfigure}
    \hfill
    \begin{subfigure}{.32\linewidth}
        \centering
        \includegraphics[width=\textwidth]{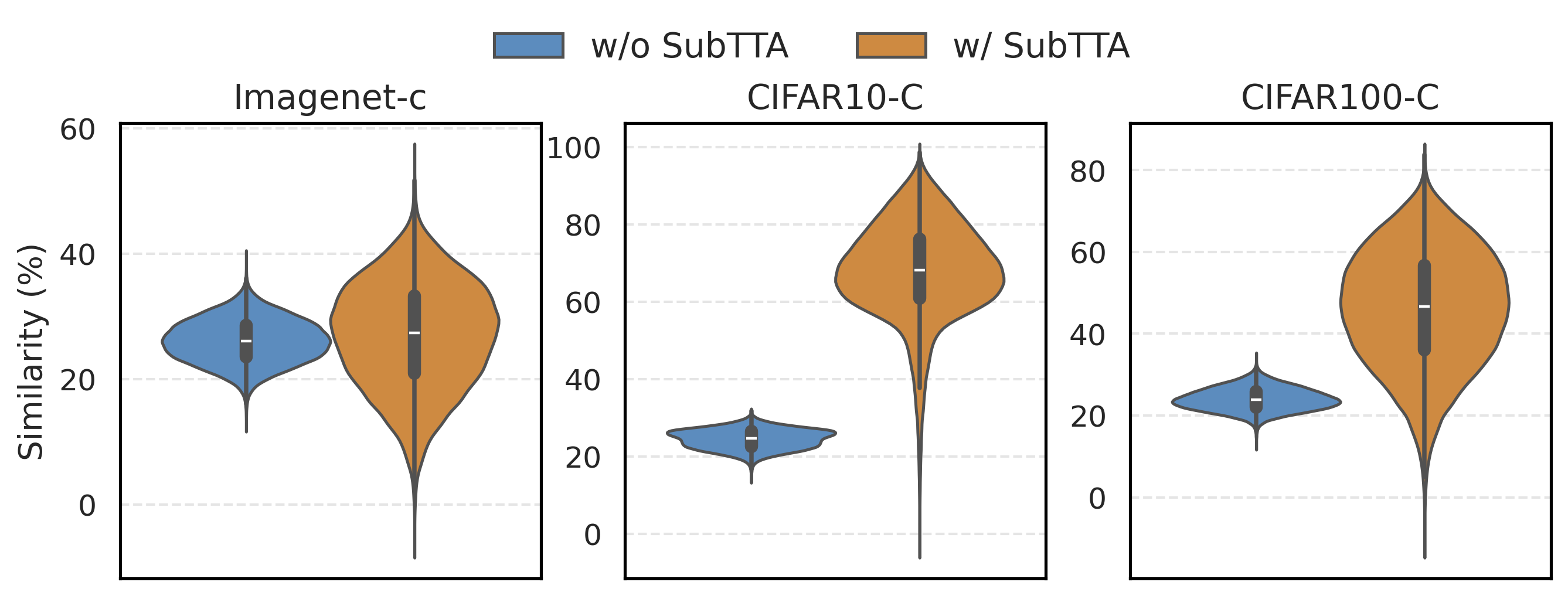}
        \caption{Visual-textual similarity.}
        \label{fig:exp-sim}
    \end{subfigure}
    \caption{Study on improving zero-shot predictions. Blue denotes source \textcolor{blue}{CLIP w/o \algname} and Orange denotes \textcolor{orange}{CLIP w/ \algname}. (a) Principal angles ($\downarrow$): \algname\ mitigates modality gap; (b) Semantic concentration ($\uparrow$): \algname\ alleviates visual nuisance; (c) Visual-textual similarity ($\uparrow$): \algname\ improves pseudo-label quality.}
    \label{fig:exp-imp}
\end{figure*}

First, \algname\ acts as a universal performance booster, irrespective of the adaptation objective. 
When integrated with diverse TTA strategies, \algname\ consistently yields performance improvements. 
This indicates that \algname\ effectively rectifies the underlying structure before standard TTA is applied.

Second, we observe more pronounced performance gains on lower-capacity models (e.g., ViT-B-16 and ViT-B-32) compared to stronger ones (e.g., ViT-L-14). 
We attribute this to the fact that weaker models are inherently more susceptible to modality gap and distribution shifts. 
\algname\ effectively compensates for these intrinsic representational deficits, providing critical robustness where the pre-trained alignment is most fragile.

\begin{figure}[t]
    \centering
    \begin{subfigure}{.24\linewidth}
        \centering
        \includegraphics[width=\textwidth]{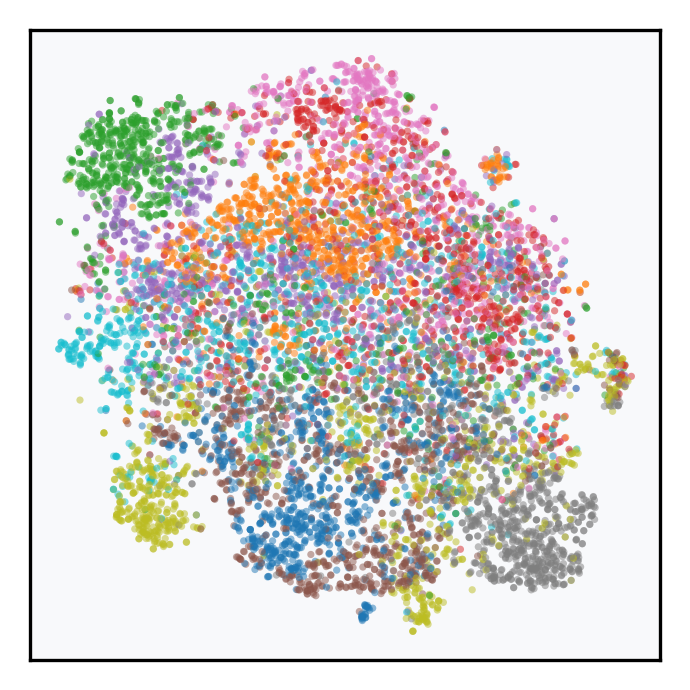}
        \caption{Source}
    \end{subfigure}
    \begin{subfigure}{.24\linewidth}
        \centering
        \includegraphics[width=\textwidth]{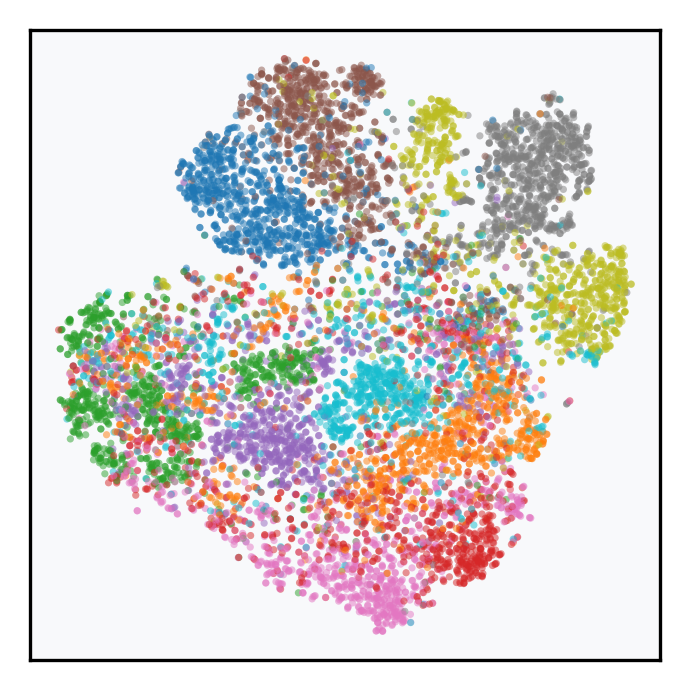}
        \caption{TENT}
    \end{subfigure}
    \begin{subfigure}{.24\linewidth}
        \centering
        \includegraphics[width=\textwidth]{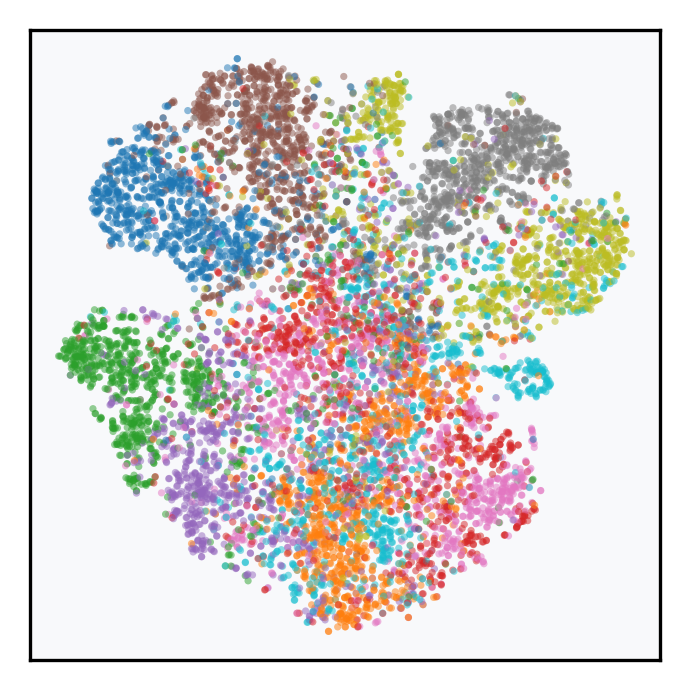}
        \caption{BATCLIP}
    \end{subfigure}
    \begin{subfigure}{.24\linewidth}
        \centering
        \includegraphics[width=\textwidth]{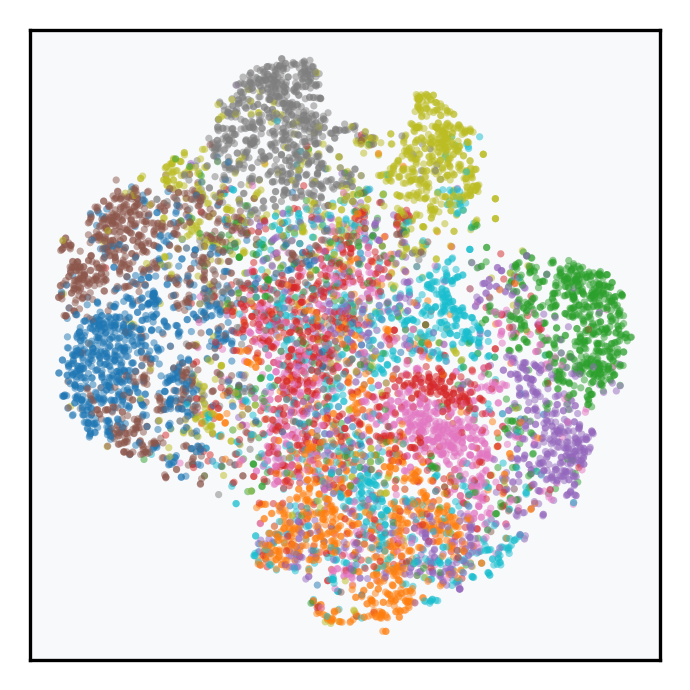}
        \caption{MINT}
    \end{subfigure}
    \begin{subfigure}{.24\linewidth}
        \centering
        \includegraphics[width=\textwidth]{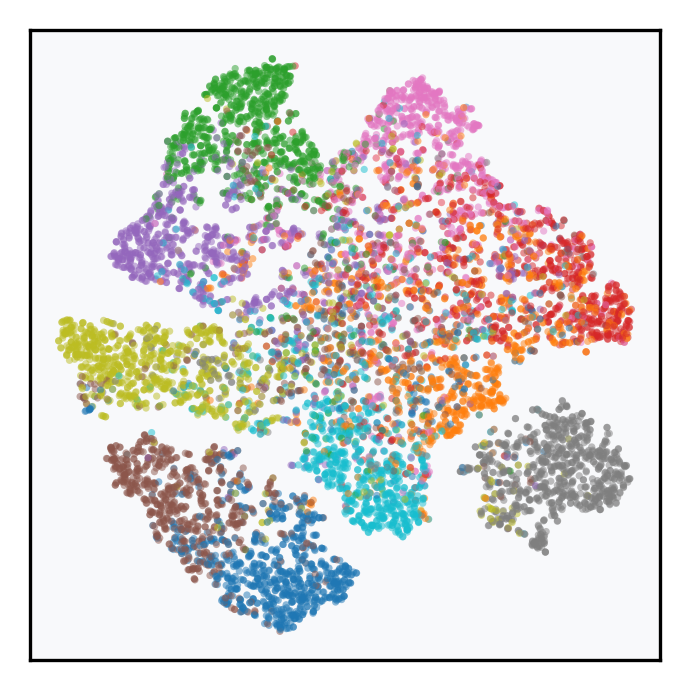}
        \caption{\algname-S}
    \end{subfigure}
    \begin{subfigure}{.24\linewidth}
        \centering
        \includegraphics[width=\textwidth]{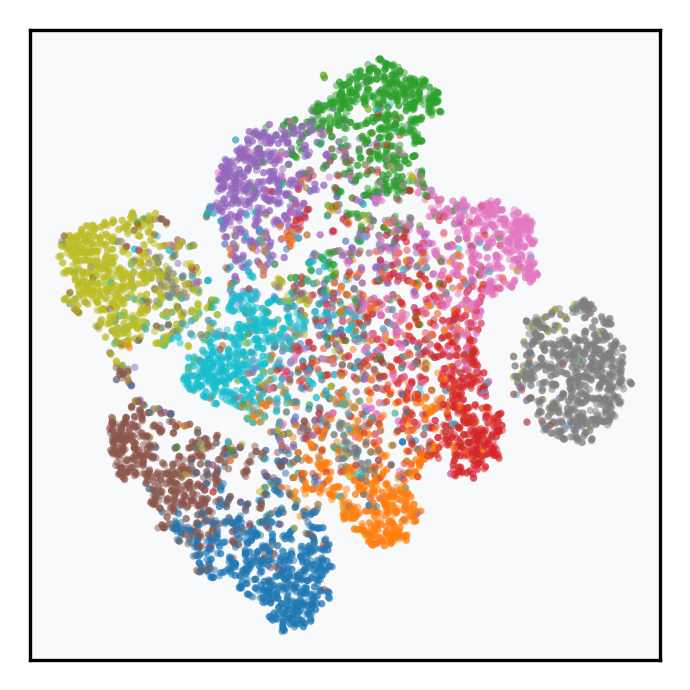}
        \caption{\algname-T}
    \end{subfigure}
    \begin{subfigure}{.24\linewidth}
        \centering
        \includegraphics[width=\textwidth]{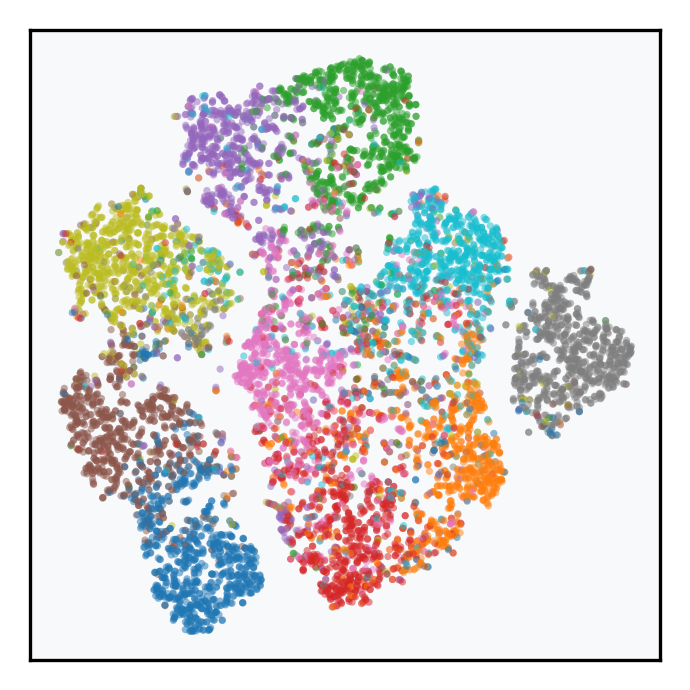}
        \caption{\algname-B}
    \end{subfigure}
    \begin{subfigure}{.24\linewidth}
        \centering
        \includegraphics[width=\textwidth]{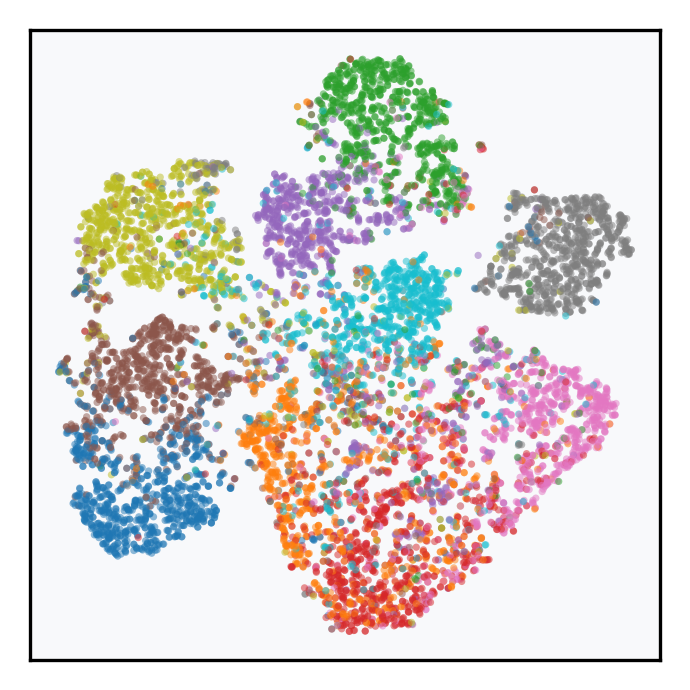}
        \caption{\algname-M}
    \end{subfigure}
    \caption{CIFAR-10-C embedding visualization of different TTA methods w/o (a-d) and w/ (e-h) \algname.}
    \label{fig:embed_vis}
\end{figure}

\subsection{On Improving Zero-shot Prediction}\label{sec:exp-zs}

In this section, we empirically validate the mechanisms through which \algname\ enhances the zero-shot prediction capability of CLIP. 
Figure~\ref{fig:exp-imp} visualizes the distributions of three key metrics, including principal angle, semantic concentration, and visual-textual similarity.
Comparing the baseline CLIP model with our \algname-augmented version, we draw the following observations.

\textbf{(1) \algname\ rectifies modality gap.}
We first examine the modality gap by analyzing the principal angles between the visual and textual subspaces.
As shown in Figure~\ref{fig:exp-angle}, CLIP without \algname\ exhibits relatively large principal angles, indicating that distribution shifts induce a significant misalignment between the visual features and their corresponding textual anchors.
In contrast, applying \algname\ leads to a marked reduction in principal angles across all datasets.
Such reduction demonstrates that \algname\ effectively rectifies the modality gap, pulling the drifted visual manifold back into geometric alignment with the invariant textual semantic space, thereby ensuring more reliable cross-modal matching.

\textbf{(2) \algname\ alleviates visual nuisance.}
We then investigate visual nuisance by measuring semantic concentration, the ratio of visual energy preserved within the task-specific textual subspace.
As shown in Figure~\ref{fig:exp-nuisance}, CLIP without \algname\ exhibits a distribution skewed toward the lower end of the semantic concentration spectrum. 
This indicates that in the raw feature space, the task-relevant semantic signal is largely overwhelmed by orthogonal components, which correspond to task-irrelevant nuisances such as background clutter or domain-specific styles. Conversely, \algname\ propels the entire distribution toward the high-concentration regime, demonstrating that the adapted features are better aligned with the textual semantic span.

\textbf{(3) \algname\ improves pseudo-label quality.}
We also assess pseudo-label quality by analyzing the cosine similarity between visual embeddings and their corresponding ground-truth textual prompts.
As shown in Figure~\ref{fig:exp-sim}, CLIP without \algname\ suffers from low visual-textual similarity scores, reflecting weak confidence in the correct semantic alignment. 
However, \algname\ significantly shifts the similarities toward higher values.
This increase indicates that the rectified visual features possess much higher semantic fidelity to their true labels.
Consequently, the zero-shot predictions derived from these enhanced similarities are not only more accurate but also more confident, providing high-quality pseudo-labels that are critical for preventing error accumulation during test-time adaptation.

\textbf{(4) \algname\ enhances embedding quality.}
In addition, we visualize the learned representations of CIFAR-10-C with and without \algname\ using t-SNE, and the results are shown in Figure~\ref{fig:embed_vis}.
As shown in the top row of Figure~\ref{fig:embed_vis}, the embeddings extracted by baselines are severely entangled, exhibiting blurred decision boundaries and cluster overlap.
This indicates a failure to disentangle task-relevant semantics from shift-induced noise, resulting in compromised discriminative power.
In contrast, integrating \algname\ (bottom row) induces a profound geometric rectification. 
First, \algname\ significantly improves \textit{intra-class compactness}, where dispersed features are pulled tighter together. 
Second, \algname\ enhances \textit{inter-class separability}, effectively pushing apart overlapping manifolds to form clear margins between categories.
This visualization validates our motivation: by projecting features onto the aligned subspace, \algname\ effectively filters out task-irrelevant nuisances and restores a discriminative semantic structure.

\input{sections/tab_episodic}
\subsection{Studies}\label{sec:exp-study}
We conduct studies to evaluate key components, analyze hyperparameter sensitivity, and assess the performance of \algname\ under episodic setting.

\subsubsection{Generalization to Episodic TTA}
Although primarily designed for online batched TTA~\cite{maharana2025batclip,bao2025mint,karmanov2024efficient,xiaodynaprompt}, \algname\ can be generalized to the episodic TTA setting, where adaptation is performed per single test sample without batch context.
By maintaining a lightweight memory bank~\cite{zhang2024dual,li2025efficient} of recent test features, we estimate a local visual subspace for geometric alignment without batch statistics, and results are shown in Table~\ref{tab:episodic}.

As observed, \algname\ safely enhances the model without suffering from the catastrophic negative adaptation that frequently plagues standard baseline methods (e.g., TENT) in the episodic setting.
This demonstrates the robustness and flexibility of our geometric alignment approach for challenging single-sample adaptation scenarios.

\subsubsection{Visual-Textual Alignment Quality}
\begin{SCfigure}
    \centering
    \includegraphics[width=0.5\columnwidth]{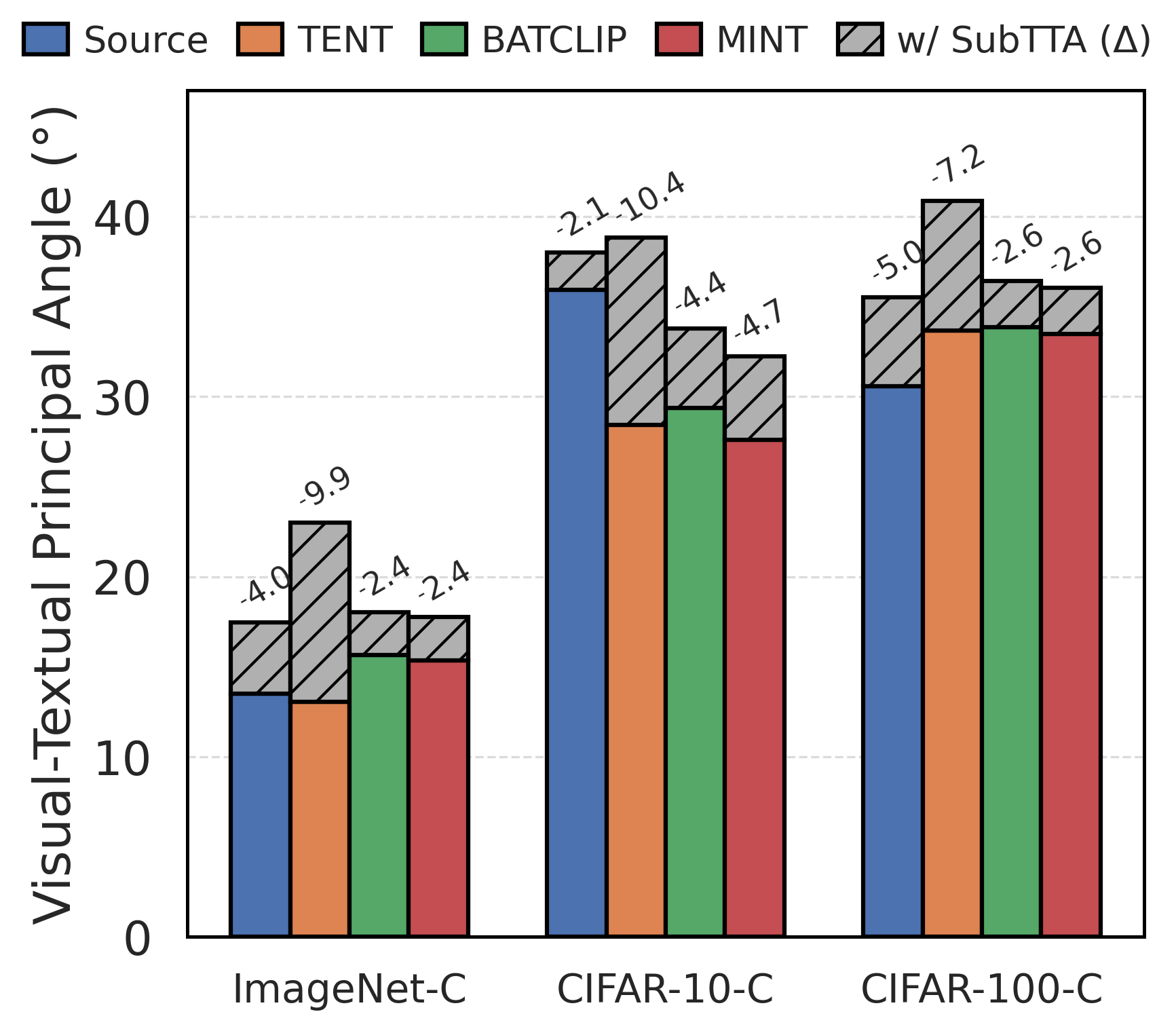}
    \caption{Visual-textual principle angles w/ (colored) and w/o (hatched) \algname. \algname\ consistently reduces the principle angles with different baselines, achieving better cross-modal alignment.}
    \label{fig:study_angle}
\end{SCfigure}

We provide a quantitative analysis on how \algname\ improves the visual-text alignment quality.
Specifically, we adopt the principal angle to measure the global geometric divergence between the entire covariance structures, i.e., $\text{span}(\mathbf{B}_\V)$ and $\text{span}(\mathbf{B}_\T)$. 
A reduction in this global metric proves that the entire visual manifold has been fundamentally rotated back into alignment with the textual semantic space.
We exam the visual-textual principal angles with and without \algname, and the result is shown in Figure~\ref{fig:study_angle}.
It is shown that \algname\ consistently reduces the visual-textual principle angles across different baselines.
Existing approaches, e.g., entropy-minimization~\cite{wang2020tent} and cluster-based loss~\cite{maharana2025batclip,bao2025mint}, mainly optimize instance-level distributions by pulling similar samples closer together, but they do not explicitly align the global geometry of the two modalities.
In contrast, \algname\ addresses this limitation by minimizing the chordal distance, which directly captures subspace-level geometric alignment between modalities, and thus serves as an effective complement to existing approaches.

\subsubsection{Ablation Study}

\begin{figure}[t]
    \centering
    \includegraphics[width=\linewidth]{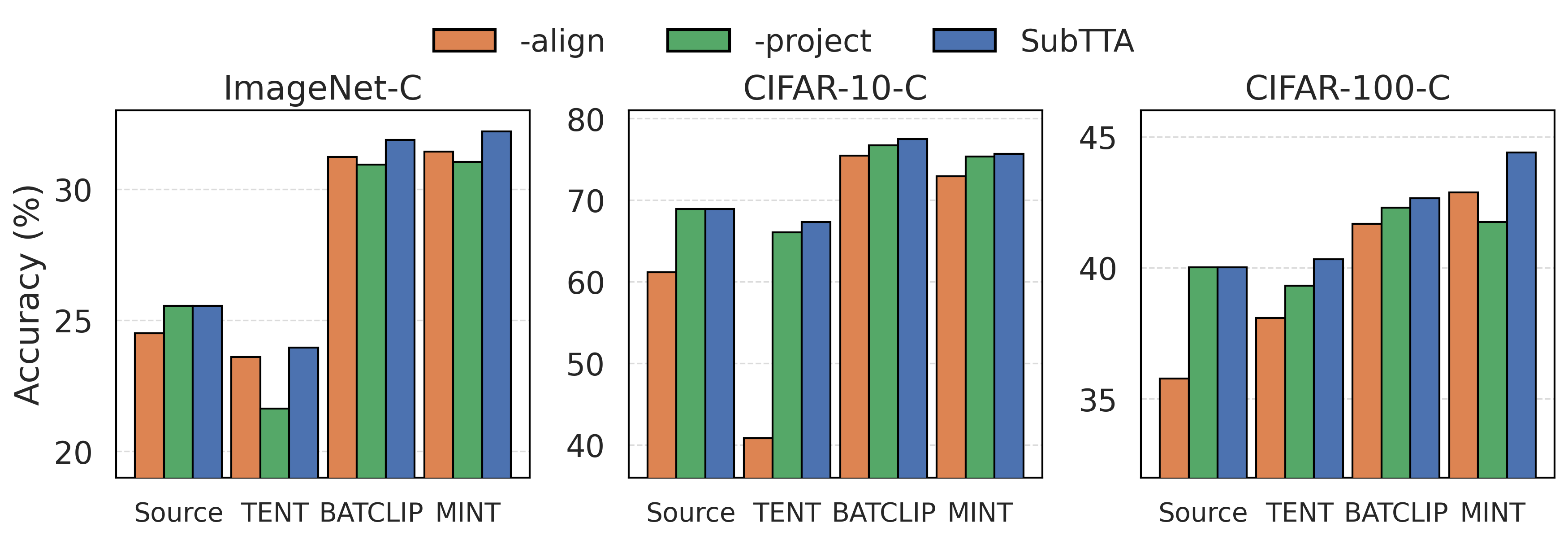}
    \caption{Ablation study on alignment and projection.}
    \label{fig:exp-ablate}
\end{figure}

We investigate the contributions of geometric alignment and semantic projection in Figure~\ref{fig:exp-ablate}, denoting their ablated versions as \texttt{-align} and \texttt{-project}, respectively.
We draw the following observations.

\noindent\textbf{(1) Alignment is a prerequisite for projection.}
The significant drop in \texttt{-align} indicates that projection is only valid when subspaces are aligned; otherwise, features may be mapped to incorrect semantic bases. 
Notably, TENT suffers from negative transfer w/o alignment, as naive projection creates confidently-wrong predictions that mislead entropy minimization.

\noindent\textbf{(2) Projection is a semantic denoiser.} 
The performance gap between \texttt{-project} and \algname\ confirms that raw embeddings contain task-irrelevant nuisances. 
Semantic projection effectively filters these dimensions, constructing a purified space that prevents optimization on spurious correlations.

\noindent\textbf{(3) Synergy of components.} 
\algname\ consistently outperforms both variants, confirming neither is redundant. 
Alignment corrects global modality drift while projection filters local visual nuisances; both are indispensable for optimal adaptation.

\subsubsection{Hyperparameter Study}\label{sec:exp-hyper}

\begin{figure}[t]
    \centering
    \includegraphics[width=\linewidth, trim=10 0 15 0, clip]{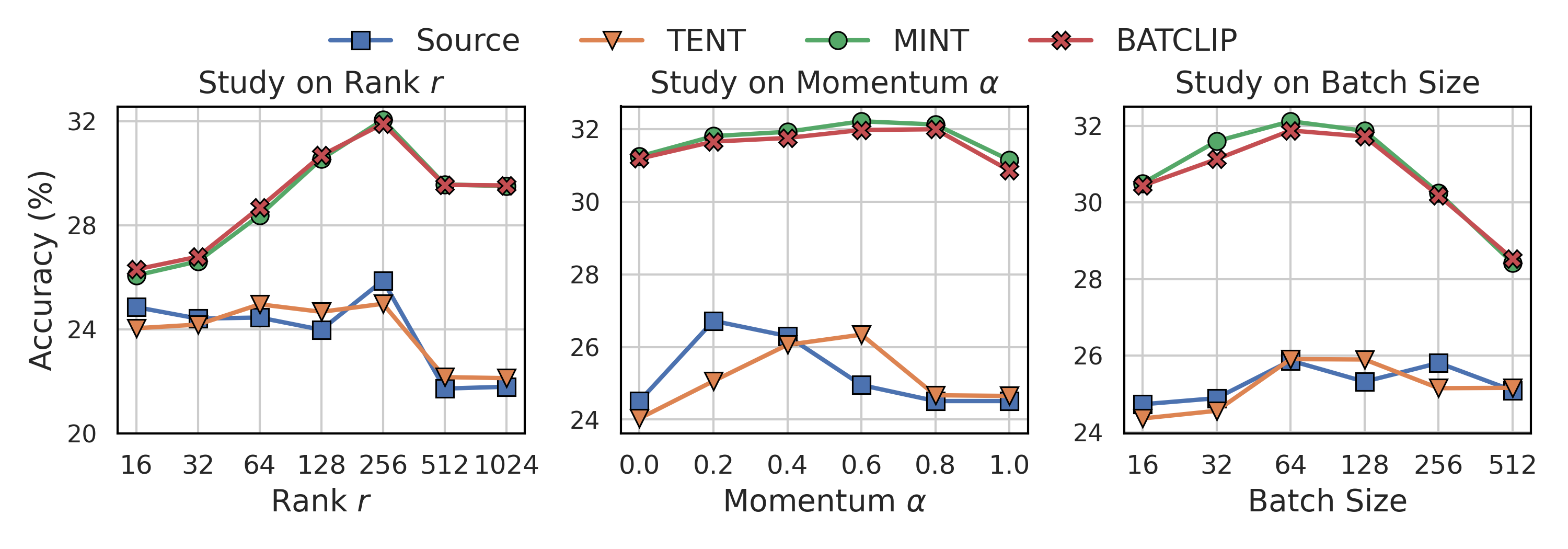}
    \caption{Hyper-parameter study. Different curves represent \algname\ with different TTA objectives.}
    \label{fig:exp-hyper}
\end{figure}

We investigate the sensitivity of \algname\ to three key hyperparameters: the subspace rank $r$, the momentum coefficient $\alpha$ and batch size.
The results are shown in Figure~\ref{fig:exp-hyper}.

\noindent\textbf{Impact of subspace rank $r$.}
When subspace rank $r$ increases, the accuracy first increases then decreases.
With small rank (e.g., $r < 64$), the performance is suboptimal because the subspace is overly compressed, causing the loss of critical semantic information.
Conversely, performance degradation is observed when $r$ approaches the full feature dimension (i.e., $d=512$ for ViT-B-16).
We attribute this to the fact that a full-rank subspace preserves all trailing principal components, hence task-irrelevant nuisances are not filtered out, which negates the denoising benefits of our subspace approach.

\noindent\textbf{Impact of momentum coefficient $\alpha$.}
Results show that \algname\ is generally robust for $\alpha \in (0.2, 0.8)$. However, performance drops significantly at the extremes.
When $\alpha=0$, the covariance estimation relies entirely on the current mini-batch. This introduces instability due to the high variance of statistics estimated from small batches, failing to capture the global visual distribution.
When $\alpha=1$, the visual covariance is fixed to the initial textual prior $\mathbf{\Sigma}_{\mathcal{T}}$ and never updates with visual features. In this scenario, the subspace alignment mechanism is effectively disabled, leading to a marked performance drop.
This confirms the necessity of our momentum-based update strategy.

\noindent\textbf{Impact of batch size.}
As batch size increases, adaptation performance initially improves then degrades at large scales (e.g., $B \ge 256$).
At small batch sizes (e.g., $B = 16$), unreliable visual covariance estimates cause subspace alignment to fluctuate erratically due to local noise rather than true manifold geometry.
While increasing the batch size stabilizes this estimation, excessively large batches suffer from an optimization deficit. 
Specifically, under a fixed-length test stream, a larger batch size results in fewer total gradient updates, thereby under-optimizing the adaptation objective.
This phenomenon is empirically validated in Figure~\ref{fig:exp-hyper}, as batch-dependent objectives (e.g., MINT, BATCLIP) degrade noticeably under fewer updates, whereas sample-wise objectives (e.g., TENT) remain relatively stable.

%% file: sections/tab_benchmark_16.tex
\begin{table*}[t]
\centering
\scriptsize
\caption{Benchmark results with ViT-B-16. We denote Top-\textcolor{blue}{1}/\textcolor{orange}{2}/\textcolor{red}{3} by Blue/Yellow/Red, respectively.}
\label{tab:benchmark-16}
\setlength{\tabcolsep}{3pt}
\renewcommand{\arraystretch}{0.95}
\resizebox{\textwidth}{!}{
\begin{tabular}{@{}ll*{15}{r}r@{}}
\toprule
\multicolumn{2}{l}{\multirow{2}{*}{\textbf{Method}}} & \multicolumn{3}{c}{Noise} & \multicolumn{4}{c}{Blur} & \multicolumn{4}{c}{Weather} & \multicolumn{4}{c}{Digital} & \multicolumn{1}{c}{\multirow{2}{*}{Mean}} \\
\cmidrule(lr){3-5}\cmidrule(lr){6-9}\cmidrule(lr){10-13}\cmidrule(lr){14-17}
\multicolumn{2}{l}{} 
& \multicolumn{1}{c}{Gauss.} 
& \multicolumn{1}{c}{Shot} 
& \multicolumn{1}{c}{Impul.}
& \multicolumn{1}{c}{Defoc.}
& \multicolumn{1}{c}{Glass}
& \multicolumn{1}{c}{Motion}
& \multicolumn{1}{c}{Zoom}
& \multicolumn{1}{c}{Snow}
& \multicolumn{1}{c}{Frost}
& \multicolumn{1}{c}{Fog}
& \multicolumn{1}{c}{Brit.}
& \multicolumn{1}{c}{Contr.}
& \multicolumn{1}{c}{Elast.}
& \multicolumn{1}{c}{Pixel.}
& \multicolumn{1}{c}{JPEG}
& \multicolumn{1}{c}{} \\
\midrule
\multirow{14}{*}{\rotatebox{90}{\textbf{ImageNet-C}}} & Source & 11.14 & 12.54 & 12.06 & 23.34 & 15.26 & 24.44 & 22.72 & 32.36 & 29.86 & 35.94 & 54.08 & 17.26 & 12.70 & 31.00 & 33.34 & 24.54 \\
 & TDA & 12.42 & 14.80 & 14.80 & 23.70 & 16.70 & 26.06 & 23.98 & 33.54 & \redcell{32.10} & 38.00 & 55.28 & 19.32 & 14.18 & 33.46 & 34.98 & 26.22 \\
 & DMN & 11.50 & 14.14 & 14.00 & 22.32 & 15.78 & 24.56 & 22.54 & 31.52 & 30.36 & 35.34 & 54.24 & 16.30 & 12.48 & 32.52 & 24.79 & 24.16 \\
 & VTE & 9.18 & 10.80 & 10.80 & 24.70 & 14.36 & 24.28 & 25.20 & \yellowcell{35.40} & \yellowcell{32.44} & 38.20 & \yellowcell{55.52} & 16.20 & 14.34 & \bluecell{38.68} & 34.04 & 25.61 \\
 & ZERO & 10.50 & 11.50 & 11.12 & 24.78 & 14.88 & 24.36 & 23.28 & 33.04 & 30.40 & 37.06 & 54.52 & 16.90 & 13.40 & 34.90 & 32.90 & 24.90 \\
 & ECALP & 13.34 & 15.58 & 14.10 & 22.56 & 15.56 & 25.52 & 23.64 & 30.96 & 30.02 & 35.78 & 52.08 & 18.34 & 13.52 & 33.14 & 33.30 & 25.16 \\
 & TENT & 5.20	& 5.68	& 7.42	& 25.30	& 19.34	& 26.86	& 24.16	& 33.52	& 30.54	& 37.80	& 54.32	& 22.52	& 13.90	& 35.08	& 36.02	& 25.18 \\
 & RoTTA & 11.44 & 13.08 & 12.36 & 23.46 & 15.48 & 24.66 & 22.78 & 32.60 & 29.92 & 35.86 & 54.18 & 17.34 & 12.92 & 31.02 & 33.52 & 24.71 \\
 & TPT & 8.54 & 9.44 & 10.16 & 24.06 & 15.06 & 25.00 & 24.04 & 33.96 & 32.18 & 37.08 & \yellowcell{55.52} & 16.52 & 13.68 & 33.90 & 33.56 & 24.85 \\
 & MEMO & 11.08 & 12.38 & 12.02 & 23.54 & 15.44 & 24.60 & 22.90 & 32.54 & 30.06 & 35.94 & 54.02 & 17.46 & 12.80 & 31.24 & 33.40 & 24.63 \\
 & WATT & 11.18 & 12.78 & 13.02 & 25.12 & 18.08 & 26.90 & 25.22 & 33.42 & 30.22 & 37.40 & 53.96 & 21.84 & 15.16 & 33.46 & 34.94 & 26.18 \\
 & STS & 15.06 & 16.84 &15.82 & 24.40 & 16.32 & 26.90 & 24.36 & 34.06 & 31.06 & 36.98 & 54.68 & \yellowcell{26.16} & 14.06 & \redcell{37.22} & 32.04 & 27.06 \\
 & MINT & \yellowcell{19.68} & \redcell{20.36} & \yellowcell{19.44} & \bluecell{26.88} & \yellowcell{21.52} & \redcell{29.88} & \redcell{25.96} & 32.68 & 29.38 & \redcell{38.84} & 55.10 & 24.30 & \redcell{19.06} & 36.38 & \bluecell{38.24} & \redcell{29.18} \\
 & BATCLIP & \redcell{19.48} & \yellowcell{21.02} & \redcell{19.30} & \yellowcell{25.92} & \redcell{21.26} & \yellowcell{29.96} & \yellowcell{28.52} & \redcell{35.22} & 31.38 & \yellowcell{40.24} & 55.20 & \redcell{25.68} & \yellowcell{23.72} & 36.82 & \redcell{37.22} & \yellowcell{30.06} \\
 & \algname & \bluecell{21.16} & \bluecell{22.88} & \bluecell{21.44} & \redcell{25.34} & \bluecell{24.82} & \bluecell{30.70} & \bluecell{29.64} & \bluecell{37.24} & \bluecell{35.90} & \bluecell{41.26} & \bluecell{56.96} & \bluecell{29.74} & \bluecell{29.40} & \yellowcell{37.58} & \yellowcell{38.18} & \bluecell{32.15} \\
\midrule
\multirow{14}{*}{\rotatebox{90}{\textbf{CIFAR10-C}}} & Source & 37.94 & 41.68 & 54.44 & 71.73 & 40.87 & 67.90 & 73.67 & 73.88 & 77.34 & 70.25 & 84.41 & 62.41 & 53.81 & 47.66 & 59.45 & 61.16 \\
 & TDA & 40.91 & 44.31 & 50.45 & 72.80 & 44.38 & 71.60 & 75.97 & 75.29 & 77.84 & 71.75 & 86.07 & 62.34 & 56.34 & 47.77 & 57.68 & 62.37 \\
 & DMN & 43.34 & 47.52 & 54.11 & 73.43 & 39.99 & 72.25 & 75.97 & 74.31 & 76.46 & 71.24 & 84.54 & 60.56 & 54.15 & 47.20 & 57.14 & 62.15 \\
 & VTE & 42.34 & 46.22 & 64.25 & 71.14 & 45.63 & 68.51 & 73.66 & 76.74 & 78.27 & 71.06 & 85.27 & 57.24 & 59.62 & 60.59 & 61.89 & 64.16 \\
 & ZERO & 39.22 & 43.77 & 57.19 & 71.91 & 40.71 & 69.22 & 74.11 & 74.36 & 77.24 & 72.44 & 83.85 & 60.74 & 55.64 & 48.97 & 62.29 & 62.11 \\
 & ECALP & 46.52 & 50.81 & 61.04 & 71.83 & 41.52 & 70.48 & 75.38 & 75.82 & 78.99 & 71.24 & 86.03 & 60.70 & 58.06 & 49.78 & 61.17 & 63.96 \\
 & TENT & 15.42	& 18.30	& 38.18	& \yellowcell{81.43}	& 21.54	& 76.33	& \yellowcell{82.24}	& \yellowcell{83.59}	& \redcell{82.24}	& \yellowcell{80.56}	& \redcell{89.76}	& \redcell{80.65}	& \redcell{63.50}	& 58.83	& 56.27	& 61.92 \\
 & RoTTA & 39.20 & 42.64 & 55.33 & 72.10 & 41.20 & 68.11 & 74.02 & 74.39 & 78.02 & 70.78 & 84.77 & 63.09 & 54.61 & 49.44 & 60.12 & 61.85 \\
 & TPT & 37.76 & 42.19 & 60.66 & 72.84 & 44.82 & 69.72 & 75.37 & 75.95 & 78.90 & 72.15 & 85.67 & 62.04 & 58.86 & 55.18 & 62.58 & 63.65 \\
 & MEMO & 37.61 & 41.26 & 55.64 & 72.08 & 41.58 & 68.60 & 73.97 & 74.96 & 77.34 & 71.50 & 84.79 & 62.07 & 55.50 & 49.23 & 61.02 & 61.81 \\
 & WATT & 45.98 & 53.25 & 60.31 & 74.99 & 38.58 & 71.49 & 75.92 & 77.69 & 80.23 & 76.33 & 87.56 & 75.59 & 55.42 & \redcell{62.04} & 63.21 & 66.57 \\
 & STS & 42.17 & 53.94 & \redcell{65.26} & 72.82 & 43.93 & 75.75 & 75.37 & 74.90 & 79.00 & 73.74 & 87.87 & 72.91 & 59.38 & 60.16 & 62.67 & 66.66 \\
 & MINT & \redcell{54.40} & \redcell{58.73} & 64.53 & 76.54 & \redcell{48.90} & \redcell{77.89} & 79.27 & 81.58 & 81.51 & 77.20 & \yellowcell{89.94} & 74.54 & 62.11 & 61.42 & \redcell{64.45} & \redcell{70.20} \\
 & BATCLIP & \bluecell{61.89} & \yellowcell{65.44} & \yellowcell{67.07} & \redcell{80.06} & \yellowcell{55.47} & \yellowcell{80.02} & \redcell{81.60} & \redcell{82.47} & \yellowcell{83.69} & 80.43 & 88.47 & \yellowcell{81.19} & \yellowcell{69.12} & \yellowcell{62.67} & \yellowcell{67.29} & \yellowcell{73.79} \\
 & \algname & \yellowcell{60.18} & \bluecell{66.62} & \bluecell{69.73} & \bluecell{81.47} & \bluecell{59.94} & \bluecell{81.24} & \bluecell{82.39} & \bluecell{83.72} & \bluecell{83.75} & \bluecell{83.48} & \bluecell{91.42} & \bluecell{86.26} & \bluecell{70.94} & \bluecell{74.42} & \bluecell{69.41} & \bluecell{76.33} \\
\midrule
\multirow{14}{*}{\rotatebox{90}{\textbf{CIFAR100-C}}} & Source & 19.57 & 21.41 & 25.26 & 42.46 & 20.05 & 43.15 & 47.92 & 48.38 & 49.67 & 41.61 & 57.01 & 34.54 & 29.23 & 23.95 & 32.46 & 35.78 \\
 & TDA & 22.59 & 25.12 & 29.22 & 43.11 & 19.35 & 43.47 & 49.27 & 48.46 & 50.40 & 41.45 & 57.96 & 35.02 & 28.73 & 24.16 & 32.46 & 36.72 \\
 & DMN & 22.02 & 25.06 & 25.43 & 40.93 & 16.10 & 43.70 & 49.38 & 45.72 & 47.88 & 39.77 & 57.61 & 32.23 & 25.89 & 22.90 & 30.99 & 35.04 \\
 & VTE & 17.97 & 18.80 & 28.22 & 40.42 & 19.58 & 39.58 & 45.36 & 48.16 & 46.85 & 40.68 & 55.30 & 30.08 & 32.47 & 30.35 & 31.52 & 35.02 \\
 & ZERO & 18.91 & 21.07 & 28.57 & 43.97 & 19.49 & 43.31 & 48.69 & 49.11 & 50.07 & 44.02 & 57.62 & 34.51 & 31.08 & 24.80 & 34.04 & 36.62 \\
 & ECALP & 23.15 & 25.17 & 30.10 & 43.19 & 19.65 & 42.82 & 49.70 & 47.97 & 50.02 & 42.31 & 58.42 & 34.63 & 30.05 & 25.19 & 32.33 & 36.98 \\
 & TENT & 7.57	& 8.23	& 8.32	& \bluecell{51.70}	& 8.17	& \bluecell{52.50}	& 53.28	& \redcell{52.16}	& 36.34	& 47.91	& 62.58	& \bluecell{52.55}	& \yellowcell{36.39}	& \yellowcell{39.98}	& \yellowcell{38.12}	& 36.98 \\
 & RoTTA & 20.62 & 22.21 & 26.28 & 42.49 & 20.28 & 43.20 & 48.10 & 48.59 & 50.00 & 41.71 & 57.24 & 34.54 & 29.18 & 25.06 & 32.93 & 36.16 \\
 & TPT & 17.86 & 19.50 & 27.15 & 43.52 & 20.02 & 42.64 & 48.66 & 49.10 & 49.49 & 42.21 & 57.29 & 33.38 & 31.08 & 27.60 & 32.79 & 36.15 \\
 & MEMO & 20.20 & 21.89 & 27.16 & 44.09 & 19.58 & 43.92 & 49.14 & 50.58 & 50.44 & 43.90 & 58.55 & 34.57 & 30.45 & 25.22 & 34.20 & 36.93 \\
 & WATT & \redcell{25.55} & 27.34 & 31.46 & 48.54 & 23.27 & 48.44 & 53.08 & 52.85 & \yellowcell{52.28} & 48.08 & 62.52 & 45.11 & \redcell{35.53} & \redcell{36.74} & \redcell{37.84} & 41.91 \\
 & STS & 25.43 & 26.20 & 29.04 & 44.01 & 24.38 & 44.05 & 48.23 & 50.62 & 50.76 & 43.50 & 60.08 & 43.90 & 33.60 & 30.59 & 34.96 & 39.29 \\
 & MINT & \yellowcell{27.49} & \yellowcell{30.46} & \yellowcell{35.97} & 49.33 & \yellowcell{26.54} & 47.37 & \redcell{53.42} & \yellowcell{52.58} & \redcell{52.10} & \bluecell{48.46} & \bluecell{64.68} & 44.15 & 35.25 & 33.02 & 35.40 & \yellowcell{42.41} \\
 & BATCLIP & 25.52 & \redcell{28.35} & \redcell{34.60} & \yellowcell{49.66} & \redcell{26.46} & \redcell{48.81} & \bluecell{54.77} & 51.90 & 51.60 & \yellowcell{48.30} & \redcell{62.77} & \redcell{45.68} & 34.82 & 33.01 & 37.25 & \redcell{42.23} \\
 & \algname & \bluecell{31.85} & \bluecell{34.25} & \bluecell{36.31} & \redcell{49.48} & \bluecell{29.66} & \yellowcell{48.86} & \yellowcell{53.86} & \bluecell{53.62} & \bluecell{54.14} & \redcell{48.19} & \yellowcell{63.20} & \yellowcell{46.47} & \bluecell{36.52} & \bluecell{40.20} & \bluecell{40.78} & \bluecell{44.49} \\
\bottomrule
\end{tabular}
}
\end{table*}

%% file: sections/tab_imagenet.tex
\begin{table}[htbp]
\centering
\scriptsize
\caption{Results on ImageNet with ViT-B-16. \textcolor{blue}{Positive}/ \textcolor{red}{negative} changes are denoted by Blue/Red, respectively.}
\label{tab:imagenet}
\setlength{\tabcolsep}{1.2pt}
\renewcommand{\arraystretch}{0.95}
\resizebox{\linewidth}{!}{
\begin{tabular}{l ccc ccc ccc}
\toprule
\multirow{2}{*}{\textbf{Method}} & \multicolumn{3}{c}{\textbf{ImageNet-R}} & \multicolumn{3}{c}{\textbf{ImageNet-K}} & \multicolumn{3}{c}{\textbf{ImageNet-A}} \\
\cmidrule(lr){2-4} \cmidrule(lr){5-7} \cmidrule(lr){8-10}
 & Vanilla & \algname & $\Delta$ & Vanilla & \algname & $\Delta$ & Vanilla & \algname & $\Delta$ \\
\midrule
Source  & 73.97 & 74.73 & \bluecell{+0.76} & 46.14 & 45.96 & \redcell{-0.18} & 47.77 & 48.07 & \bluecell{+0.30} \\
TENT    & 74.50 & 74.77 & \bluecell{+0.27} & 46.87 & 46.94 & \bluecell{+0.07} & 48.25 & 48.24 & \redcell{-0.01} \\
MINT    & 74.38 & 75.92 & \bluecell{+1.54} & 47.75 & 48.53 & \bluecell{+0.78} & 48.39 & 49.14 & \bluecell{+0.75} \\
BATCLIP & 74.42 & 75.65 & \bluecell{+1.23} & 47.07 & 47.53 & \bluecell{+0.46} & 48.35 & 49.08 & \bluecell{+0.73} \\
\bottomrule
\end{tabular}
}
\end{table}

%% file: sections/tab_episodic.tex
\begin{table}[t]
\small
\centering
\caption{Performance under episodic TTA.}
\label{tab:episodic}
\setlength{\tabcolsep}{3.5pt}
\begin{tabular}{lcccccc}
\toprule
\multirow{2}{*}{\textbf{Method}} & \multicolumn{2}{c}{\textbf{ImageNet-C}} & \multicolumn{2}{c}{\textbf{CIFAR10-C}} & \multicolumn{2}{c}{\textbf{CIFAR100-C}} \\
\cmidrule(lr){2-3} \cmidrule(lr){4-5} \cmidrule(lr){6-7}
 & Source & TENT & Source & TENT & Source & TENT \\
\midrule
Vanilla & 24.54 & 24.53 & 61.17 & 61.16 & 35.79 & 35.79 \\
\algname  & 24.68 & 24.74 & 61.49 & 61.95 & 36.24 & 36.45 \\
\midrule
$\Delta$ & \bluecell{+0.14} & \bluecell{+0.21} & \bluecell{+0.32} & \bluecell{+0.79} & \bluecell{+0.45} & \bluecell{+0.66} \\
\bottomrule
\end{tabular}
\end{table}

%% file: sections/5-con.tex
\section{Conclusion}
In this work, we address the vulnerability of VLMs to distribution shifts by investigating the prevalent yet precarious reliance on raw zero-shot predictions.
We identify that adaptation failures stem from modality gap and visual nuisance within the shifted embedding space.
To overcome these barriers, we propose \algname, a novel framework that shifts the TTA from noisy self-training to robust subspace geometry. 
By explicitly aligning the visual subspace to the textual semantic anchor via chordal distance and projecting features onto a purified task-specific subspace, \algname\ effectively rectifies modality gap and alleviates visual nuisance.
Extensive experiments validate the effectiveness of \algname\ in enhancing various TTA methods.

\section*{Acknowledgement}
This work is supported by NSF (2416070) and DARPA (HR0011263E079).
The content of the information in this document does not necessarily reflect the position or the policy of the Government, and no official endorsement should be inferred.  The U.S. Government is authorized to reproduce and distribute reprints for Government purposes notwithstanding any copyright notation here on.

\section*{Limitations}
While \algname\ demonstrates robust performance, we acknowledge a few limitations. 
First, our method relies on batch-level statistics to estimate reliable subspaces. Consequently, it is not immediately applicable to the strictly episodic setting, where model is adapted to a single test sample, without a mechanism to accumulate historical samples (e.g., a memory bank). 
Second, as an optimization-based approach, \algname\ requires an open-source model with access to model gradients, restricting its usage with closed-source proprietary APIs. 
Lastly, the SVD operation introduces a slight computational overhead compared to training-free baselines. 
Future work could explore efficient approximations for eigendecomposition to further reduce latency, or extend the subspace alignment principle to black-box adaptation scenarios.

%% file: sections/app.tex
\newpage
\appendix
\section*{Appendix}

\section{Algorithm}\label{sec:app-algo}
We provide the pseudo code of \algname\ in Algorithm~\ref{alg:subtta}, which includes three steps: geometric alignment, semantic projection and standard TTA.
\input{sections/algo}

\section{Further Discussions}
\subsection{Generalization to Episodic TTA}
The default configuration of \algname\ operates under the standard online TTA paradigm, where data arrives in continuous batches. 
However, in strictly episodic TTA, where adaptation is performed on single isolated test samples, or extreme small-batch scenarios, relying solely on current statistics becomes infeasible. 
The lack of sufficient batch context leads to severe high-variance or rank-deficient covariance estimation, which subsequently collapses the visual subspace.

To elegantly extend \algname\ to these challenging regimes, we seamlessly integrate a lightweight memory bank mechanism~\cite{zhang2024dual,li2025efficient}. 
Specifically, we maintain a First-In-First-Out (FIFO) queue of size $M$ to dynamically accumulate the most recent test features. 
Instead of computing the visual subspace based purely on the current, inadequate observation, we compute the empirical covariance over the buffered features in the memory bank. 
This mechanism effectively smooths the statistical fluctuations and constructs a reliable local representation of the data distribution.
Consequently, even in the complete absence of large batch statistics, the memory bank ensures a smooth and stable statistical subspace estimation, providing a robust foundation for \algname's cross-modal geometric alignment.

\subsection{Textual Subspace as Semantic Anchors}
We follow the standard practice in existing VLM TTA approaches~\cite{maharana2025batclip,bao2025mint,deng2025panda,osowiechi2024watt} by updating the normalization layers of the visual encoder while keeping the text encoder and prompts frozen. 
The rationale for utilizing the textual subspace as a semantic anchor stems from the inherent differences between the two modalities: textual prompts (e.g., \texttt{“a photo of a <class>”}) encode highly distilled, task-relevant semantics.
Conversely, image embeddings often entangle the core object with task-irrelevant visual nuisances, such as background, lighting, and stylistic variations.

However, relying exclusively on the textual subspace assumes it perfectly encapsulates the task semantics.
To mitigate the potential risks of a biased or imperfect textual subspace, two potential strategies can be integrated into our framework:
\begin{itemize}[noitemsep, topsep=0pt]
    \item \textbf{Prompt Ensembling}: Similar to~\cite{osowiechi2024watt}, by employing multiple diverse prompt templates, we can construct a more robust ensembled textual subspace. This approach broadens the semantic coverage and significantly reduces the model's sensitivity to the biases of any single prompt.
    \item \textbf{Soft Projection via Residual Connections}: Instead of enforcing a rigid projection in Eq.~\eqref{eq:proj}, we may introduce a residual connection as follows
    \begin{equation*}
        \text{proj}(\mathbf{v})=\mathbf{v}\mathbf{B}_T^\top\mathbf{B}_T+\lambda\mathbf{v},
    \end{equation*}
    where $\lambda$ is a hyperparameter governing the weight of the residual signal. This soft projection formulation deliberately preserves a fraction of the features outside the textual subspace, thereby preventing the excessive discarding of potentially useful visual cues when the textual anchor is imperfect.
\end{itemize}

\section{Experiments}\label{app:exp}


\subsection{Results with Different Backbones}
To verify the scalability and generalization of our approach, we provide detailed benchmark results using ViT-B-32 and ViT-L-14 backbones in Table~\ref{tab:benchmark-32} and Table~\ref{tab:benchmark-14}, respectively. The results align with the observations on ViT-B-16 reported in the main text, confirming three key trends:

\input{sections/tab_benchmark_32}
\input{sections/tab_benchmark_14}

\noindent\textbf{(1) Consistent SOTA performance across model capacities.} \algname\ maintains its superiority regardless of the backbone architecture. On the lower-capacity ViT-B-32, \algname\ achieves a mean accuracy of 30.50\% on ImageNet-C, outperforming the runner-up \textsc{BATCLIP} (27.37\%) by a substantial margin of 3.13\%. On the stronger ViT-L-14, \algname\ continues to set the state-of-the-art with 44.55\% mean accuracy. This confirms that our geometric rectification is effective for both rescuing weaker models and refining stronger ones.

\noindent\textbf{(2) Elimination of negative adaptation.} Similar to the ViT-B-16 settings, baseline methods exhibit instability under heavy noise. For instance, on ViT-B-32, \textsc{TENT} suffers from negative adaptation on Gaussian Noise (12.16\% vs. Source 12.90\%). In contrast, \algname\ consistently improves over the source baseline across all corruption categories for both backbones, demonstrating robust stability against severe distribution shifts.

\noindent\textbf{(3) Validation of the paradigm hierarchy.} The performance hierarchy observed in the main text holds true across different backbones: training-based methods generally outperform training-free ones, and cluster-based objectives (e.g., \textsc{MINT}, \textsc{BATCLIP}) consistently surpass entropy-based approaches (e.g., \textsc{TENT}). \algname\ builds upon the robust cluster-based paradigm and further elevates it via subspace alignment, yielding the most reliable adaptation performance.

\subsection{Robustness against Shift Levels}
\begin{figure}[t]
    \centering
    \includegraphics[width=\linewidth, trim=8 0 0 0, clip]{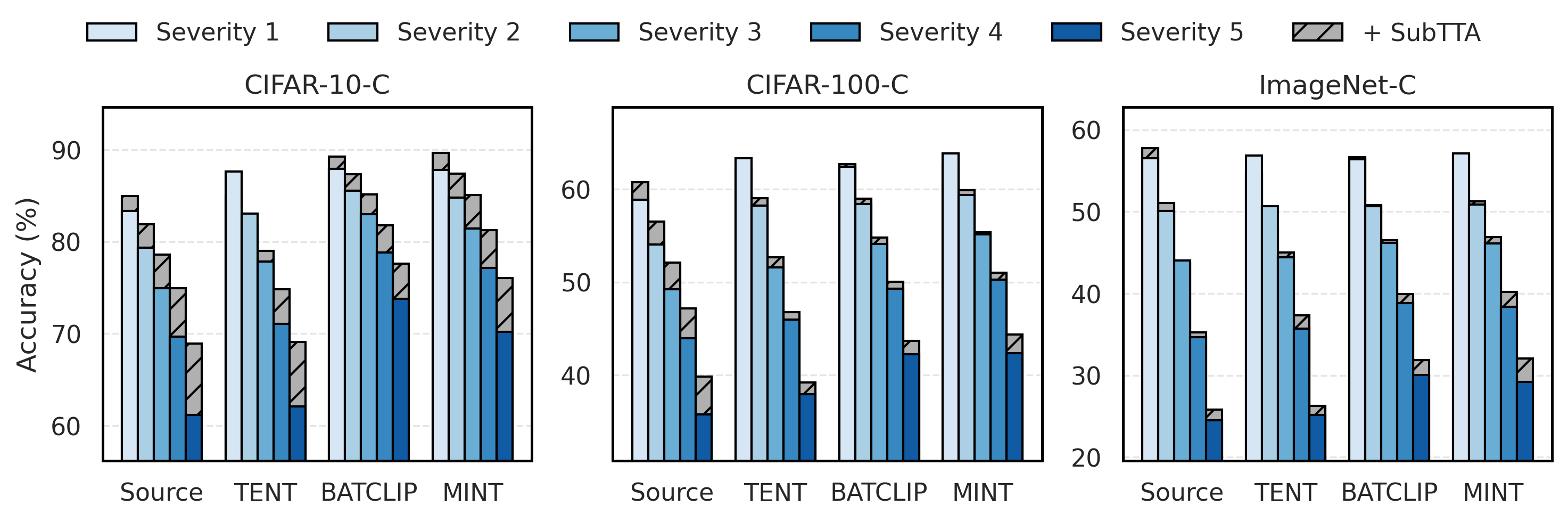}
    \vspace{-15pt}
    \caption{TTA performance under different shift levels.}
    \vspace{-15pt}
    \label{fig:severity}
\end{figure}

We evaluate how model performs under different shift levels, and the results are shown in Figure~\ref{fig:severity}.
First, we observe a consistent performance degradation in all baseline methods (colored bars) as severity increases, confirming the vulnerability of pre-trained CLIP models to severe distribution shifts. 
Second, the efficacy of \algname\ correlates positively with shift severity. 
While the improvements are consistent at lower levels, the performance boost (hatched bars) becomes more significant at higher levels.
This indicates that \algname\ is particularly critical in extreme scenarios, effectively acting as a safeguard where standard baselines struggle most.

\subsection{Effect of Exponential Moving Average}
We carry out experiments to understand the role of exponential moving average (EMA) in statistical estimation of \algname.
While an EMA may theoretically reduce the immediate responsiveness of test-time adaptation, it is a crucial component in \algname\ for ensuring smooth and reliable subspace estimation. 
Under challenging conditions, such as extremely small batch sizes or datasets with a massive number of classes, a single batch of test samples may fail to cover the global data distribution comprehensively. 
Relying exclusively on such isolated batches can result in ill-conditioned or heavily biased subspace representations, which significantly degrades overall performance.

To empirically validate this, we conduct an ablation study by removing the EMA module. Specifically, we set the update rate to $\alpha = 1$ in Eq.~\eqref{eq:ema}, forcing the model to rely solely on the current batch statistics for subspace computation.
As demonstrated in Table~\ref{tab:momentum}, discarding EMA leads to a consistent and noticeable performance drop across benchmarks, confirming that historical statistical aggregation is indispensable.

Fundamentally, the update hyperparameter $\alpha$ governs the trade-off between responsiveness (favoring the current batch) and robustness (leveraging historical context). 
Based on empirical observations, an appropriate $\alpha$ can be selected based on the following principles:
\begin{itemize}[noitemsep, topsep=0pt]
    \item Favor a smaller $\alpha$ (More Robustness): In scenarios characterized by small batch sizes, a large number of classes, or relatively stationary test distributions, a smaller $\alpha$ is preferred. This retains more historical information, ensuring stable statistical estimation despite high per-batch variance.
    \item Favor a larger $\alpha$ (More Responsiveness): Conversely, in settings with large batch sizes, a compact label space, or rapidly changing, non-stationary test streams, a larger $\alpha$ allows the model to swiftly adapt to new distributions by relying more on recent batch statistics.
\end{itemize}
\input{sections/tab_momentum}

\subsection{Sensitivity to Prompt Design}
To understand \algname's reliance on the textual anchor, e.g., when textual prompts are noisy or incomplete, we conduct experiments to analyze \algname's sensitivity to different prompt designs.
Specifically, we consider the following 8 prompts, and evaluate VLM performance with and without \algname.
\begin{tcolorbox}
[left=3pt,right=3pt,top=3pt,bottom=3pt,colback=gray!5!white,colframe=black!75!white,title=Prompt Templates]
    \textbf{T0}: A photo of a \{class\}\\
    \textbf{T1}: itap of a \{class\}\\
    \textbf{T2}: a bad photo of the \{class\}\\
    \textbf{T3}: a origami \{class\}\\
    \textbf{T4}: a photo of the large \{class\}\\
    \textbf{T5}: a \{class\} in a video game\\
    \textbf{T6}: art of the \{class\}\\
    \textbf{T7}: a photo of the small \{class\}
\end{tcolorbox}
\input{sections/tab_prompt}
As the results in Table~\ref{tab:prompt} shows, \algname\ consistently enhances baseline performance across all datasets and prompt templates, with average improvement of 1.16\% on ImageNet-C, 6.39\% on CIFAR10-C and 4.06\% on CIFAR100-C.
Depite strong stylistic or domain biases of different prompts, \algname\ maintains steady improvements even under these conditions. 
This indicates that as long as the prompt spans a valid semantic category, \algname\ can successfully filter out orthogonal visual corruptions regardless of the stylistic anchor.

\subsection{Sensitivity to the Number of Classes}
The number of classes in the target domain plays a critical role in the quality of subspace estimation~\cite{zanella2025realistic}, as a larger semantic space inherently introduces greater complexity and potential for misaligned feature projections. 
To investigate the robustness of our approach, we evaluate \algname\ on ImageNet-C by sampling subsets of classes, with the total number of classes varying from 100 to 1,000.

As illustrated in Table~\ref{tab:num_class}, we observe two key findings.
First, \algname\ exhibits remarkable robustness to the expansion of the label space.
While the absolute accuracy of all methods naturally decays as the classification task becomes more challenging (with \#classes from 100 to 1,000), the performance of \algname\ scales gracefully without catastrophic degradation. 
Second, and more importantly, \algname\ enhances the performance of the baseline TTA method across all class regimes in most cases (38 out of 40). 
Notably, the performance gap between MINT equipped with \algname\ and the vanilla MINT baseline remains stable, and even becomes more pronounced, at larger class counts (e.g., 500 to 1,000 classes).
This demonstrates that our geometric filtering mechanism effectively purifies representations even when the semantic space is highly dense and complex.

\input{sections/tab_num_class}

\subsection{Latency Analysis}
We provide a latency analysis of the proposed \algname.
Theoretically, let $B$ denote the batch size, $d$ the feature dimension, $r$ the subspace rank, and $C$ the number of classes. 
The dominant computational overhead of \algname\ arises from the subspace computation (Eq.~\eqref{eq:basis}), which is bounded by $O(d^2r)$. 
In contrast, baseline TTA methods, e.g., MINT and BATCLIP, the cluster-based TTA loss incurs a computational complexity of $O(BCd)$.

Under practical deployment scenarios, these two complexities are of comparable magnitude. 
For typical VLM configurations, the hidden dimension is $d \in [512, 1024]$, the subspace rank is $r \ll d$, the batch size is $B \in [64, 256]$, and the number of classes $C$ ranges from hundreds to over a thousand. 
Crucially, the complexity of \algname\ is independent of the number of classes $C$, whereas cluster-based TTA methods scale linearly with $C$. 
This property renders \algname\ highly scalable and particularly well-suited for settings with a large vocabulary space.

Empirically, we evaluate the runtime efficiency across the three benchmark datasets under default configurations.
Table~\ref{tab:latency} reports the per-batch latency (in seconds) attributed to the subspace operations ($t_{\text{Sub}}$) and the subsequent adaptation process ($t_{\text{TTA}}$).
As demonstrated,$t_{\text{Sub}}$ and $t_{\text{TTA}}$ are comparable, indicating that the subspace projection does not introduce a substantial per-batch latency increase.
These findings confirm that the geometric filtering mechanism in \algname\ imposes a marginal computational burden, maintaining overall efficiency relative to existing TTA routines.

\input{sections/tab_latency}

\section{Datasets}\label{sec:app-dataset}

We first introduce three corrupted datasets, including ImageNet-C, CIFAR10-C, and CIFAR100-C.
\textbf{\texttt{ImageNet-C}}~\cite{hendrycks2019benchmarking} is a robustness benchmark derived from ImageNet. It spans 1,000 object classes and contains 50,000 test images for each corruption type and severity level.
\textbf{\texttt{CIFAR-10-C}}~\cite{hendrycks2019benchmarking} is constructed from the CIFAR-10. It comprises 10,000 images distributed across 10 distinct classes.
\textbf{\texttt{CIFAR-100-C}}~\cite{hendrycks2019benchmarking} is an extension of the CIFAR-100, containing 10,000 images covering 100 fine-grained classes.

All three datasets employ a shared set of 15 algorithmically generated corruptions to assess model robustness under distribution shifts. These corruptions are categorized into four primary groups: noise, blur, weather, and digital distortions, and each is applied at five severity levels.
The \textit{noise} category includes Gaussian, shot, and impulse noise, which introduce random pixel-level variations.
The \textit{blur} category encompasses defocus, glass, motion, and zoom blur, simulating various optical distortions.
\textit{Weather}-related corruptions, such as snow, frost, and fog, replicate environmental conditions that obscure image details.
Lastly, \textit{digital} distortions include brightness, contrast, elastic transform, pixelate, and JPEG compression, reflecting common post-processing or compression artifacts.

Besides, we also consider three datasets with realistic perturbations, including ImageNet-A, ImageNet-R and ImageNet-K.
\textbf{\texttt{ImageNet-A}}~\cite{hendrycks2021natural} is a challenging out-of-distribution benchmark containing natural, unperturbed images that commonly cause classification errors in standard ResNet models. It spans 200 object classes sampled from the original ImageNet and consists of 7,500 test images.
\textbf{\texttt{ImageNet-R}}~\cite{hendrycks2021many} is constructed to evaluate model robustness to abstract visual renditions. It comprises around 30,000 images distributed across 200 distinct ImageNet classes.
\textbf{\texttt{ImageNet-K}}~\cite{wang2019learning} is a domain shift benchmark constructed from sketch-like images. It contains 50,000 images covering the full 1,000 fine-grained classes of the original ImageNet.

\section{More Related Works}
\label{app:related_works}

We provide more related works on pre-trained foundation models, domain adaptation and cross-domain alignment. 

\paragraph{Pre-trained Foundation Models.}
Foundation models have reshaped the field by pretraining models on vast amounts of web-scale data~\cite{achiam2023gpt,touvron2023llama,lin2025quantization,ai2025resmoe,cui2026adafuse,xu2026prune}.
Large Language Models (LLMs) have demonstrated remarkable generalization capabilities with various applications in natural language understanding~\cite{touvron2023llama,zhang2025improving, lin2026efficient}, numerical analysis~\cite{li2026language,jing2026tsaqa,lin2026mixture}, and ethics evaluation~\cite{lin2025moralise,lin2026alert}.
Parallel to this success, encoder-decoder VLMs such as CLIP~\cite{radford2021learning} and ALIGN~\cite{jia2021scaling} extend this paradigm to the visual domain by aligning images and text in a shared embedding space via contrastive learning. 
This joint training enables powerful zero-shot transfer to downstream tasks without task-specific fine-tuning. 
Decoder-only VLMs like Flamingo~\cite{alayrac2022flamingo}, BLIP-2~\cite{li2023blip}, and LLaVA~\cite{liu2023visual} leverage visual instruction tuning to project visual features into the input space of frozen LLMs, unlocking capabilities for complex multimodal understanding and dialogue.
Despite their impressive capabilities, these models remain vulnerable to distribution shifts when deployed in open-world environments, and significant efforts have been made on improving the robustness of LLMs and VLMs~\cite{qi2024visual,wang2023decodingtrust,yan2025answer,zeng2025harnessing,lin2024duquant,lin2026duquant++}, necessitating effective adaptation strategies to bridge the gap between pre-training and testing stages.

\paragraph{Domain Adaptation.}
The challenge of generalizing models to out-of-distribution data has evolved through increasingly constrained settings, spanning from label-scarce scenarios to fully unsupervised and source-free environments.

\emph{(Semi-)supervised domain adaptation (SSDA)} represents the most accessible settings, assuming the availability of at least a few labeled samples in the target domain~\cite{motiian2017unified,saito2019semi}. 
Approaches typically leverage the limited target labels to perform fine-tuning or align class-conditional distributions via metric learning~\cite{kang2019contrastive} and minimax entropy training~\cite{saito2019semi}.
However, the reliance on target annotation, even if minimal, limits their scalability in open-world deployments.

\emph{Unsupervised Domain Adaptation (UDA)} removes the reliance on target labels, assuming access to both labeled source data and fully unlabeled target data.
Methods in this realm can be broadly categorized into two streams.
One prominent line of works employ adversarial learning~\cite{ganin2015unsupervised}, optimizing a domain discriminator to force the feature extractor to learn domain-invariant representations.
Another line of works minimize the discrepancy between source and target distributions such as maximum mean discrepancy~\cite{long2015learning,yan2017mind}, Wasserstein distance~\cite{shen2018wasserstein,zeng2025pave} and correlation alignment~\cite{sun2016deep,sun2017correlation,lin2025cats}.

\emph{Test-time adaptation (TTA)} faces the most challenging setting, imposing strict latency and online constraints where data arrives in streams~\cite{wang2020tent,yu2026inference}. 
Existing TTA strategies generally diverge into three optimization categories:
First (entropy minimization~\cite{wang2020tent,zhang2022memo}), which sharpens prediction distributions to reduce uncertainty; 
Second (self-training with pseudo-labels~\cite{rusak2022if,bao2024matcha,yang2024simce}), which utilizes robust loss functions or cluster structures to refine decision boundaries.
Third (parameter-efficient tuning~\cite{shu2022test}), which updates only specific modules (e.g., prompts or normalization layers) to prevent catastrophic forgetting.

\paragraph{Cross-Domain Alignment.}
To mitigate the discrepancy between domains, various alignment strategies~\cite{yan2021dynamic,yan2021bright,zeng2023parrot,zeng2024hierarchical,yu2025joint,yu2025planetalign} have been proposed to learn domain-invariant representations~\cite{wang2018acekg,wang2023networked,wang2023noisy}, with applications in various data modalities such as image~\cite{zhu2019adapting,xu2020cross,yu2026avatar}, text~\cite{liu2021bapa,chen2020graph}, graphs~\cite{yan2023trainable,yan2023reconciling,yan2024pacer,yan2024thegcn,xu2024slog,zeng2023generative,zeng2024graph,chen2026tag}, and recommendation~\cite{zhao2023cross,zhao2023crossdomain,zeng2025interformer,zeng2025hierarchical,yoo2024ensuring,liu2024collaborative,liang2025external}.
One prevalent approach is \textit{statistical moment matching}, which explicitly minimizes the distance between feature distributions~\cite{long2015learning, sun2016deep}. 
Another significant direction is \textit{manifold alignment}, which posits that domain shifts can be modeled as geometric transformations~\cite{fernando2013unsupervised,gong2012geodesic}. 
While these classical methods laid the theoretical groundwork, they typically require offline processing or access to source data. 
\algname\ revitalizes these geometric principles, adapting the manifold alignment concept to the challenging online, source-free TTA setting by utilizing the textual subspace as a stable semantic anchor.

\input{sections/checklist}

%% file: sections/algo.tex
\begin{algorithm}[]
\caption{\algname}
\label{alg:subtta}
\begin{algorithmic}[1]
\small
\REQUIRE
    Image encoder $f_v$; text encoder $f_t$;
    image batches $\mathcal{D}_{\text{test}}=\{\mathbf{X}^{(1)},\mathbf{X}^{(2)},\dots\}$;
    textual prompts $t_1,\dots,t_C$;
    subspace rank $r$; momentum coefficient $\alpha$.
\ENSURE Prediction $\hat{Y}$ for the test images.
\AlgStep{\textit{Stage 1: Textual Subspace Initialization}}
\STATE Extract normalized text features $\mathbf{T}=[\mathbf{t}_1,\dots,\mathbf{t}_C]^\top \in\mathbb{R}^{C \times d}$ with $\mathbf{t}_i=f_t(t_i)$.
\STATE Compute textual covariance $\mathbf{\Sigma}_{\mathcal{T}} = \mathbf{T}^\top \mathbf{T}$.
\STATE Compute textual basis $\mathbf{B}_{\mathcal{T}} \in \mathbb{R}^{r \times d}$ via Eq.~\eqref{eq:basis}.
\STATE Initialize visual covariance prior $\mathbf{\Sigma}_{\mathcal{V}} \leftarrow \mathbf{\Sigma}_{\mathcal{T}}$.
\AlgStep{\textit{Stage 2: Online Test-Time Adaptation}}
\FOR{each test batch $\mathbf{X}^{(k)}$ in $\mathcal{D}_{\text{test}}$}
    \STATE Extract visual features $\mathbf{V}^{(k)} = f_v(\mathbf{X}^{(k)})\in\mathbb{R}^{B\times d}$.
    \STATE Visual covariance EMA via Eq.~\eqref{eq:ema}
    \STATE Compute visual basis $\mathbf{B}_{\mathcal{V}} \in \mathbb{R}^{r \times d}$ via Eq.~\eqref{eq:basis}.
    \AlgStep{\texttt{Step 1: Geometric Alignment}}
    \STATE Compute alignment loss $\mathcal{L}_{\text{align}}$ in Eq.~\eqref{eq:loss-align}.
    \STATE Update model parameters $\theta$ by Adam with $\nabla_\theta \mathcal{L}_{\text{align}}$.
    \AlgStep{\texttt{Step 2: Semantic Projection}}
    \STATE Extract image features $\tilde{\mathbf{V}}^{(k)}$ using updated model.
    \STATE Project features onto textual subspace via Eq.~\eqref{eq:proj}.
    \AlgStep{\texttt{Step 3: Standard TTA}}
    \STATE Compute standard TTA loss $\mathcal{L}_{\text{TTA}}$ (e.g., MINT or Entropy) on purified features $\mathbf{V}_{\text{proj}}$.
    \STATE Update model parameters $\theta$ by Adam with $\nabla_\theta \mathcal{L}_{\text{TTA}}$.
    \STATE Compute prediction $\hat{Y}^{(k)}=\argmax_c \mathbf{V}_{\text{proj}}\mathbf{T}^\top$.
\ENDFOR
\STATE \textbf{return} TTA predictions $\hat{Y}^{(1)}, \hat{Y}^{(2)}, \dots$.
\end{algorithmic}
\end{algorithm}

%% file: sections/tab_benchmark_32.tex
\begin{table*}[t]
\centering
\scriptsize
\caption{Benchmark results with ViT-B-32. We denote Top-\textcolor{blue}{1}/\textcolor{orange}{2}/\textcolor{red}{3} by Blue/Yellow/Red, respectively.}
\label{tab:benchmark-32}
\setlength{\tabcolsep}{3pt}
\renewcommand{\arraystretch}{0.95}
\resizebox{\textwidth}{!}{
\begin{tabular}{@{}ll*{15}{r}r@{}}
\toprule
\multicolumn{2}{l}{\multirow{2}{*}{\textbf{Method}}} & \multicolumn{3}{c}{Noise} & \multicolumn{4}{c}{Blur} & \multicolumn{4}{c}{Weather} & \multicolumn{4}{c}{Digital} & \multicolumn{1}{c}{\multirow{2}{*}{Mean}} \\
\cmidrule(lr){3-5}\cmidrule(lr){6-9}\cmidrule(lr){10-13}\cmidrule(lr){14-17}
\multicolumn{2}{l}{} & \multicolumn{1}{c}{Gauss.} & \multicolumn{1}{c}{Shot} & \multicolumn{1}{c}{Impul.} & \multicolumn{1}{c}{Defoc.} & \multicolumn{1}{c}{Glass} & \multicolumn{1}{c}{Motion} & \multicolumn{1}{c}{Zoom} & \multicolumn{1}{c}{Snow} & \multicolumn{1}{c}{Frost} & \multicolumn{1}{c}{Fog} & \multicolumn{1}{c}{Brit.} & \multicolumn{1}{c}{Contr.} & \multicolumn{1}{c}{Elast.} & \multicolumn{1}{c}{Pixel.} & \multicolumn{1}{c}{JPEG} & \multicolumn{1}{c}{} \\
\midrule
\multirow{14}{*}{\rotatebox[origin=c]{90}{\textbf{ImageNet-C}}} & Source & 12.90 & 13.06 & 12.80 & 24.38 & 11.82 & 22.68 & 20.24 & 25.60 & 25.84 & 30.24 & 50.48 & 17.30 & 18.98 & 32.32 & 29.10 & 23.18 \\
& TDA & 12.80 & 14.72 & 14.84 & 24.14 & 12.98 & 23.46 & 20.98 & 26.74 & \yellowcell{28.00} & 32.28 & \bluecell{51.64} & 18.16 & 20.70 & 33.00 & 30.20 & 24.31 \\
& DMN & 13.00 & 14.32 & 14.50 & 22.92 & 11.78 & 22.18 & 19.68 & 23.70 & 25.72 & 29.62 & 50.50 & 14.78 & 19.92 & 32.46 & 29.36 & 22.96 \\
& VTE & 11.98 & 12.32 & 13.44 & 25.06 & 11.60 & 22.60 & 22.34 & 27.40 & 27.04 & 32.28 & \yellowcell{51.62} & 16.82 & 20.02 & 34.80 & 32.78 & 24.14 \\
& ZERO & 12.36 & 12.88 & 12.58 & 24.90 & 12.06 & 22.84 & 21.46 & 26.72 & 26.64 & 30.58 & \redcell{50.98} & 16.80 & 19.66 & 33.82 & 30.52 & 23.65 \\
& ECALP & 14.40 & 15.10 & 14.68 & 23.28 & 12.18 & 23.18 & 21.08 & 24.40 & 25.38 & 29.50 & 47.92 & 18.26 & 19.12 & 31.42 & 29.00 & 23.26 \\
& RoTTA & 13.22 & 13.30 & 13.10 & 24.52 & 12.02 & 22.93 & 20.30 & 25.88 & 26.00 & 30.34 & 50.46 & 17.40 & 19.10 & 32.38 & 29.22 & 23.34 \\
& TPT & 12.16 & 12.60 & 12.48 & \redcell{25.36} & 12.22 & 22.42 & 20.94 & 26.70 & 26.78 & 30.50 & 50.82 & 16.88 & 19.88 & 33.40 & 30.50 & 23.58 \\
& MEMO & 12.84 & 13.00 & 12.78 & 24.66 & 11.88 & 22.82 & 20.40 & 25.70 & 26.00 & 30.28 & 50.50 & 17.32 & 19.02 & 32.48 & 29.14 & 23.25 \\
& WATT & 14.28 & 14.64 & 14.42 & \yellowcell{25.62} & 15.26 & 25.52 & 21.92 & 26.76 & 25.86 & 31.44 & 50.70 & \yellowcell{22.22} & 19.68 & 33.56 & 30.64 & 24.83 \\
& MINT & \yellowcell{18.84} & \yellowcell{19.98} & \yellowcell{19.42} & 24.86 & \redcell{19.08} & \redcell{27.20} & \redcell{22.44} & \redcell{27.94} & \redcell{27.28} & \redcell{32.72} & 50.64 & 20.18 & \redcell{24.04} & \redcell{34.84} & \redcell{33.14} & \redcell{26.84} \\
& BATCLIP & \redcell{17.64} & \redcell{18.60} & \redcell{16.98} & 24.56 & \yellowcell{20.40} & \bluecell{28.46} & \yellowcell{24.66} & \yellowcell{29.16} & 27.02 & \yellowcell{35.76} & 49.56 & \redcell{20.56} & \yellowcell{27.62} & \yellowcell{35.24} & \yellowcell{34.26} & \yellowcell{27.37} \\
& \algname & \bluecell{22.56} & \yellowcell{23.52} & \bluecell{23.02} & \bluecell{25.66} & \bluecell{23.36} & \bluecell{28.46} & \bluecell{27.42} & \bluecell{31.18} & \bluecell{32.32} & \bluecell{39.08} & 50.50 & \bluecell{26.00} & \bluecell{33.00} & \bluecell{37.12} & \bluecell{34.28} & \bluecell{30.50} \\
\midrule
\multirow{14}{*}{\rotatebox[origin=c]{90}{\textbf{CIFAR10-C}}} & Source & 35.51 & 40.01 & 43.17 & 69.91 & 41.46 & 64.52 & 70.10 & 70.84 & 72.32 & 66.65 & 81.35 & 64.48 & 59.69 & 48.18 & 56.62 & 58.99 \\
& TDA & 41.49 & 43.44 & 41.44 & 71.14 & 44.86 & 66.96 & 72.37 & 72.20 & 74.53 & 68.46 & 83.57 & 65.64 & 62.61 & 51.14 & 55.67 & 61.03 \\
& DMN & 38.83 & 39.45 & 42.12 & 70.74 & 39.48 & 65.01 & 72.78 & 72.52 & 73.41 & 66.45 & 82.39 & 58.46 & 60.51 & 49.05 & 55.92 & 59.14 \\
& VTE & 47.55 & 50.18 & \redcell{53.11} & 71.35 & 53.87 & 67.89 & 72.95 & \redcell{76.37} & \redcell{76.24} & 70.76 & 83.33 & 61.05 & \yellowcell{68.99} & 58.57 & 61.07 & 64.89 \\
& ZERO & 37.11 & 41.66 & 46.79 & 71.21 & 42.51 & 65.21 & 71.58 & 72.27 & 73.79 & 68.92 & 82.78 & 65.53 & 62.35 & 47.66 & 58.80 & 60.54 \\
& ECALP & 44.36 & 47.05 & 43.93 & 71.17 & 43.56 & 68.32 & 73.92 & 72.49 & 74.92 & 68.06 & 82.61 & 62.95 & 61.91 & 47.18 & 55.67 & 61.21 \\
& RoTTA & 36.36 & 41.12 & 43.76 & 69.94 & 42.44 & 64.68 & 70.06 & 71.34 & 72.83 & 67.33 & 81.74 & 65.12 & 60.41 & 49.37 & 57.26 & 59.58 \\
& TPT & 43.13 & 46.67 & 48.30 & 71.33 & 47.77 & 66.95 & 72.05 & 73.95 & 76.09 & 68.73 & 84.16 & 66.31 & 63.91 & 51.82 & 58.02 & 62.61 \\
& MEMO & 36.50 & 40.68 & 44.28 & 70.81 & 42.10 & 64.67 & 70.63 & 72.17 & 73.01 & 67.63 & 82.10 & 64.25 & 61.03 & 47.84 & 57.61 & 59.69 \\
& WATT & 43.63 & 48.69 & 49.13 & 72.23 & 46.15 & 66.74 & 71.04 & 73.64 & 74.46 & \redcell{71.25} & 83.99 & \redcell{73.12} & 61.98 & \yellowcell{59.25} & \yellowcell{62.98} & 63.89 \\
& MINT & \yellowcell{54.18} & \yellowcell{57.78} & 47.30 & \redcell{73.49} & \yellowcell{56.01} & \redcell{73.71} & \yellowcell{76.39} & 74.94 & 74.46 & 70.16 & \redcell{85.66} & 70.60 & 64.66 & \redcell{58.85} & 59.53 & \redcell{66.51} \\
& BATCLIP & \redcell{52.19} & \redcell{55.70} & \yellowcell{53.96} & \yellowcell{76.09} & \redcell{55.09} & \yellowcell{74.75} & \redcell{75.18} & \yellowcell{77.23} & \yellowcell{78.09} & \yellowcell{74.97} & \yellowcell{86.03} & \yellowcell{77.37} & \redcell{67.53} & 58.01 & \redcell{61.90} & \yellowcell{68.27} \\
& \algname & \bluecell{60.43} & \bluecell{63.78} &\bluecell{56.40} & \bluecell{78.48} & \bluecell{61.66} & \bluecell{76.24} & \bluecell{77.26} & \bluecell{80.03} & \bluecell{79.99} & \bluecell{77.95} & \bluecell{88.72} & \bluecell{83.62} & \bluecell{70.51} & \bluecell{73.77} & \bluecell{63.46} & \bluecell{72.82} \\
\midrule
\multirow{14}{*}{\rotatebox[origin=c]{90}{\textbf{CIFAR100-C}}} & Source & 16.18 & 17.76 & 17.55 & 39.08 & 17.63 & 38.55 & 43.81 & 42.34 & 43.41 & 39.59 & 50.38 & 29.41 & 28.80 & 22.83 & 29.38 & 31.78 \\
& TDA & 18.21 & 21.36 & 19.35 & 40.51 & 17.73 & 40.16 & 45.11 & 44.78 & 45.55 & 40.37 & 52.16 & 30.42 & 29.32 & 22.58 & 31.42 & 33.27 \\
& DMN & 17.26 & 20.32 & 13.44 & 36.86 & 15.23 & 40.23 & 45.85 & 42.77 & 44.00 & 39.05 & 51.71 & 27.21 & 27.93 & 19.54 & 29.44 & 31.39 \\
& VTE & 16.83 & 18.33 & 19.01 & 39.62 & 22.89 & 39.10 & 43.83 & 44.58 & 44.86 & 39.18 & 49.39 & 28.34 & \redcell{34.14} & 26.92 & 30.11 & 33.14 \\
& ZERO & 15.91 & 17.91 & 19.53 & 41.10 & 17.18 & 39.79 & 44.78 & 43.33 & 43.47 & 40.99 & 51.27 & 29.41 & 30.62 & 21.81 & 31.14 & 32.55 \\
& ECALP & 18.36 & 19.56 & 17.87 & 38.74 & 17.72 & 39.64 & 44.71 & 43.34 & 45.14 & 40.60 & 52.56 & 31.01 & 30.33 & 23.01 & 30.17 & 32.85 \\
& RoTTA & 16.57 & 18.30 & 17.73 & 38.58 & 17.67 & 38.58 & 43.71 & 42.44 & 43.38 & 39.25 & 50.72 & 28.96 & 28.83 & 23.43 & 29.79 & 31.86 \\
& TPT & 15.99 & 17.58 & 17.44 & 39.17 & 19.57 & 38.90 & 43.89 & 43.56 & 44.49 & 40.03 & 50.92 & 27.73 & 30.95 & 23.33 & 29.61 & 32.21 \\
& MEMO & 16.51 & 18.17 & 18.19 & 41.49 & 17.48 & 40.34 & 45.94 & 44.18 & 43.94 & 40.79 & 52.64 & 29.60 & 30.52 & 22.78 & 31.05 & 32.91 \\
& WATT & 21.40 & 22.34 & \yellowcell{23.29} & \bluecell{47.58} & 18.63 & \redcell{44.03} & \yellowcell{49.82} & \yellowcell{47.62} & \yellowcell{47.86} & \yellowcell{45.32} & \redcell{57.62} & \bluecell{43.61} & 33.05 & \yellowcell{29.71} & \bluecell{35.81} & \yellowcell{37.85} \\
& MINT & \yellowcell{23.81} & \yellowcell{26.29} & 21.36 & 45.67 & \yellowcell{24.00} & 42.27 & 49.00 & 46.20 & 44.95 & 43.11 & 56.20 & \redcell{38.88} & 34.05 & 28.08 & 32.47 & 37.09 \\
& BATCLIP & \redcell{21.45} & \redcell{24.54} & \redcell{22.85} & \redcell{46.17} & \redcell{23.20} & \bluecell{44.86} & \yellowcell{49.82} & \redcell{47.01} & \redcell{46.62} & \redcell{44.98} & \bluecell{58.58} & 38.72 & \yellowcell{34.69} & \redcell{28.45} & \redcell{33.23} & \redcell{37.68} \\
& \algname & \bluecell{25.58} & \bluecell{27.39} & \bluecell{24.28} & \yellowcell{46.73} & \bluecell{26.62} & \yellowcell{44.17} & \bluecell{50.04} & \bluecell{47.87} & \bluecell{47.90} & \bluecell{45.58} & \yellowcell{57.72} & \yellowcell{41.09} & \bluecell{37.02} & \bluecell{32.97} & \yellowcell{35.75} & \bluecell{39.38} \\
\bottomrule
\end{tabular}
}
\end{table*}

%% file: sections/tab_benchmark_14.tex
\begin{table*}[t]
\centering
\scriptsize
\caption{Benchmark results with ViT-L-14. We denote Top-\textcolor{blue}{1}/\textcolor{orange}{2}/\textcolor{red}{3} by Blue/Yellow/Red, respectively.}
\label{tab:benchmark-14}
\setlength{\tabcolsep}{3pt}
\renewcommand{\arraystretch}{0.95}
\resizebox{\textwidth}{!}{
\begin{tabular}{@{}ll*{15}{r}r@{}}
\toprule
\multicolumn{2}{l}{\multirow{2}{*}{\textbf{Method}}} & \multicolumn{3}{c}{Noise} & \multicolumn{4}{c}{Blur} & \multicolumn{4}{c}{Weather} & \multicolumn{4}{c}{Digital} & \multicolumn{1}{c}{\multirow{2}{*}{Mean}} \\
\cmidrule(lr){3-5}\cmidrule(lr){6-9}\cmidrule(lr){10-13}\cmidrule(lr){14-17}
\multicolumn{2}{l}{} & \multicolumn{1}{c}{Gauss.} & \multicolumn{1}{c}{Shot} & \multicolumn{1}{c}{Impul.} & \multicolumn{1}{c}{Defoc.} & \multicolumn{1}{c}{Glass} & \multicolumn{1}{c}{Motion} & \multicolumn{1}{c}{Zoom} & \multicolumn{1}{c}{Snow} & \multicolumn{1}{c}{Frost} & \multicolumn{1}{c}{Fog} & \multicolumn{1}{c}{Brit.} & \multicolumn{1}{c}{Contr.} & \multicolumn{1}{c}{Elast.} & \multicolumn{1}{c}{Pixel.} & \multicolumn{1}{c}{JPEG} & \multicolumn{1}{c}{} \\
\midrule
\multirow{14}{*}{\rotatebox[origin=c]{90}{\textbf{ImageNet-C}}} & Source & 27.40 & 29.42 & 28.72 & 34.58 & 25.28 & 40.96 & 36.70 & 49.80 & 44.10 & 49.78 & 65.36 & 35.18 & 30.42 & 53.48 & 42.24 & 39.56 \\
 & TDA & 27.52 & 30.72 & \redcell{32.30} & 35.20 & 27.04 & 42.60 & 38.20 & 51.40 & \yellowcell{46.36} & \redcell{52.08} & \redcell{66.76} & 37.70 & 31.76 & 55.22 & 44.00 & 41.26 \\
 & DMN & 27.44 & 30.16 & 30.84 & 32.72 & 25.98 & 41.26 & 37.00 & 49.54 & 44.56 & 50.02 & 66.00 & 29.50 & 31.10 & 54.90 & 42.18 & 39.55 \\
 & VTE & 26.36 & 29.20 & 28.98 & 36.38 & 24.88 & 40.92 & \yellowcell{39.36} & 50.52 & 44.92 & 50.86 & \yellowcell{67.42} & 33.80 & 30.56 & \yellowcell{55.58} & 47.76 & 40.50 \\
 & ZERO & 27.12 & 28.70 & 28.72 & 35.54 & 25.66 & 40.50 & 38.16 & 50.56 & 43.84 & 50.32 & 66.04 & 35.24 & 30.74 & 55.08 & 44.98 & 40.08 \\
 & ECALP & 29.48 & 30.34 & 30.86 & 35.60 & 27.50 & 42.82 & 38.38 & 50.48 & 45.42 & 50.52 & 65.40 & 37.44 & 30.12 & 55.16 & 42.80 & 40.82 \\
 & RoTTA & 27.62 & 29.60 & 29.02 & 34.68 & 25.50 & 41.18 & 36.82 & 50.04 & 44.20 & 49.76 & 65.40 & 35.08 & 30.82 & 53.64 & 42.36 & 39.71 \\
 & TPT & 27.08 & 30.00 & 29.54 & \redcell{35.84} & 25.78 & 41.50 & 37.98 & \redcell{51.64} & 45.72 & 51.78 & \bluecell{67.78} & 36.34 & 31.64 & \redcell{55.56} & 46.16 & 40.96 \\
 & MEMO & 27.42 & 29.46 & 28.74 & 34.66 & 25.34 & 41.00 & 36.74 & 49.78 & 44.16 & 49.80 & 65.42 & 35.16 & 30.40 & 53.56 & 42.24 & 39.59 \\
 & WATT & 29.82 & \redcell{31.42} & 30.70 & \yellowcell{36.68} & 28.92 & 42.30 & 38.48 & 51.10 & 44.46 & 50.70 & 65.60 & \redcell{39.18} & 33.54 & 54.16 & 43.30 & 41.36 \\
 & MINT & \yellowcell{32.00} & \yellowcell{33.20} & \yellowcell{33.84} & \bluecell{37.22} & \yellowcell{33.38} & \redcell{43.68} & \redcell{39.28} & \bluecell{53.62} & \redcell{45.62} & \yellowcell{53.42} & 66.34 & 39.16 & \bluecell{37.32} & 55.46 & \bluecell{49.22} & \yellowcell{43.52} \\
 & BATCLIP & \redcell{31.44} & 31.40 & 30.94 & \redcell{35.84} & \redcell{31.68} & \yellowcell{43.72} & 39.26 & 50.16 & 43.02 & 51.12 & 65.50 & \yellowcell{40.14} & \redcell{35.38} & 53.68 & \yellowcell{48.92} & \redcell{42.15} \\
 & \algname & \bluecell{32.96} & \bluecell{35.28} & \bluecell{36.48} & 35.70 & \bluecell{34.86} & \bluecell{44.02} & \bluecell{41.34} & \yellowcell{53.22} & \bluecell{47.12} & \bluecell{55.70} & 65.74 & \bluecell{46.30} & \yellowcell{35.88} & \bluecell{55.82} & \redcell{47.84} & \bluecell{44.55} \\
\midrule
\multirow{14}{*}{\rotatebox[origin=c]{90}{\textbf{CIFAR10-C}}} & Source & 64.33 & 67.10 & 75.90 & 80.47 & 51.55 & 80.34 & 83.04 & 83.21 & 84.92 & 79.14 & 91.07 & 84.09 & 66.44 & 73.60 & 72.26 & 75.83 \\
 & TDA & 68.32 & 70.62 & 75.97 & 80.67 & 50.18 & 81.03 & 83.07 & 84.32 & 85.82 & 79.02 & 91.78 & 84.65 & 66.94 & 73.93 & 73.70 & 76.67 \\
 & DMN & 68.64 & 69.94 & 74.76 & 76.33 & 51.12 & 79.33 & 80.86 & 84.94 & 85.87 & 79.15 & 91.94 & 84.57 & 66.20 & 71.97 & 73.34 & 75.93 \\
 & VTE & 63.90 & 66.97 & 75.24 & 78.72 & 51.65 & 79.43 & 81.32 & 83.60 & 85.27 & 76.43 & 90.13 & 81.33 & 70.11 & 74.85 & 70.79 & 75.32 \\
 & ZERO & 66.12 & 69.49 & 76.85 & 81.97 & 53.45 & 80.71 & 84.40 & 85.19 & 86.59 & 79.83 & 92.19 & 84.51 & 69.94 & 74.40 & 73.71 & 77.29 \\
 & ECALP & 70.41 & 72.13 & \redcell{79.90} & 80.93 & 54.42 & 81.07 & 83.70 & 84.92 & 86.51 & 79.44 & 92.25 & 84.69 & 68.12 & 74.67 & 73.61 & 77.78 \\
 & RoTTA & 65.73 & 68.49 & 76.57 & 80.54 & 52.39 & 80.51 & 83.29 & 83.56 & 85.35 & 79.51 & 91.28 & 84.30 & 67.01 & 74.62 & 73.07 & 76.41 \\
 & TPT & 67.15 & 69.87 & 79.01 & 80.83 & 55.33 & 81.17 & 83.43 & 85.14 & 86.97 & 79.81 & 92.16 & 83.87 & 70.87 & 76.52 & 74.07 & 77.75 \\
 & MEMO & 64.88 & 67.74 & 76.40 & 80.83 & 52.16 & 80.58 & 83.46 & 83.84 & 85.51 & 79.40 & 91.38 & 84.11 & 67.45 & 73.90 & 73.20 & 76.32 \\
 & WATT & \redcell{70.84} & 72.44 & 77.03 & 82.83 & 61.91 & 82.35 & 85.15 & 85.70 & 86.98 & 81.93 & 91.84 & 88.25 & 72.04 & 77.76 & \redcell{77.15} & 79.61 \\
 & MINT & 68.12 & \redcell{73.40} & 77.52 & \redcell{85.58} & \redcell{63.27} & \redcell{83.93} & \redcell{87.00} & \redcell{86.77} & \redcell{88.39} & \redcell{83.15} & \yellowcell{94.12} & \redcell{88.96} & \redcell{73.42} & \redcell{79.10} & 76.07 & \redcell{80.59} \\
 & BATCLIP & \yellowcell{74.81} & \yellowcell{78.11} & \yellowcell{83.70} & \yellowcell{87.53} & \bluecell{71.62} & \yellowcell{86.74} & \bluecell{89.29} & \bluecell{89.99} & \yellowcell{89.73} & \yellowcell{87.53} & \bluecell{94.65} & \yellowcell{92.29} & \bluecell{78.92} & \yellowcell{83.37} & \bluecell{80.09} & \yellowcell{84.56} \\
 & \algname & \bluecell{77.28} & \bluecell{79.97} & \bluecell{84.14} & \bluecell{87.67} & \yellowcell{70.00} & \bluecell{86.87} & \yellowcell{88.82} & \yellowcell{88.95} & \bluecell{89.87} & \bluecell{88.28} & \redcell{93.81} & \bluecell{93.37} & \yellowcell{78.42} & \bluecell{83.86} & \yellowcell{78.90} & \bluecell{84.68} \\
\midrule
\multirow{14}{*}{\rotatebox[origin=c]{90}{\textbf{CIFAR100-C}}} & Source & 35.52 & 37.69 & 50.30 & 50.20 & 25.98 & 52.58 & 57.15 & 54.87 & 58.85 & 49.63 & 67.13 & 53.50 & 36.43 & 44.82 & 42.79 & 47.83 \\
 & TDA & 40.09 & 42.74 & 52.27 & 50.96 & 26.13 & 53.67 & 58.55 & 56.48 & 60.12 & 50.90 & 68.78 & 55.08 & 37.23 & 46.46 & 46.37 & 49.72 \\
 & DMN & 38.07 & 41.97 & 51.01 & 46.36 & 22.71 & 53.35 & 58.83 & 50.67 & 55.05 & 50.17 & 65.81 & 52.00 & 35.73 & 45.65 & 42.54 & 47.33 \\
 & VTE & 36.31 & 39.04 & 48.97 & 50.43 & 26.06 & 51.30 & 55.67 & 58.59 & 60.31 & 47.37 & 66.88 & 51.76 & \bluecell{41.59} & \redcell{49.06} & 44.67 & 48.53 \\
 & ZERO & 35.70 & 37.36 & 52.04 & 53.51 & 26.63 & 54.08 & 58.97 & 57.54 & 60.92 & 51.68 & 69.08 & 54.92 & 39.35 & 46.04 & 45.72 & 49.57 \\
 & ECALP & 40.27 & 42.16 & 53.61 & 51.81 & 27.91 & 55.20 & 59.93 & 58.24 & \yellowcell{61.27} & 52.04 & 70.25 & 56.14 & 38.46 & 46.98 & 46.69 & 50.73 \\
 & RoTTA & 36.03 & 38.46 & 51.49 & 50.42 & 26.51 & 52.62 & 57.25 & 54.64 & 58.97 & 49.58 & 66.94 & 53.53 & 36.84 & 45.90 & 43.22 & 48.16 \\
 & TPT & 38.53 & 40.29 & 51.86 & 51.54 & 26.98 & 53.05 & 57.86 & 58.52 & 61.20 & 49.99 & 68.27 & 52.75 & 40.30 & 47.36 & 45.73 & 49.62 \\
 & MEMO & 36.28 & 38.44 & 51.69 & 51.60 & 26.67 & 53.43 & 58.14 & 56.22 & 60.04 & 50.38 & 68.19 & 53.89 & 37.61 & 45.60 & 44.23 & 48.83 \\
 & WATT & \yellowcell{42.07} & \redcell{43.50} & \redcell{53.95} & \redcell{55.36} & 30.25 & \bluecell{56.28} & \redcell{61.47} & \redcell{59.15} & \redcell{61.15} & \redcell{54.11} & 70.24 & 60.34 & 40.63 & 48.51 & \yellowcell{47.69} & \redcell{52.31} \\
 & MINT & \bluecell{43.26} & \bluecell{46.02} & \yellowcell{54.94} & \yellowcell{57.24} & \yellowcell{32.89} & \redcell{55.69} & \bluecell{62.82} & \bluecell{61.85} & 61.02 & \yellowcell{54.81} & \bluecell{73.72} & \yellowcell{62.42} & \redcell{40.73} & \bluecell{51.84} & \yellowcell{47.69} & \yellowcell{53.80} \\
 & BATCLIP & 40.11 & 43.10 & 53.47 & 54.79 & \redcell{30.58} & 53.30 & 59.77 & 57.27 & 58.60 & 50.18 & \redcell{70.44} & \redcell{60.87} & 35.94 & 47.77 & 47.17 & 50.89 \\
 & \algname & \redcell{41.36} & \yellowcell{44.54} & \bluecell{57.17} & \bluecell{57.67} & \yellowcell{34.39} & \yellowcell{55.84} & \yellowcell{62.35} & \yellowcell{60.85} & \yellowcell{61.72} & \bluecell{54.93} & \yellowcell{72.27} & \bluecell{64.54} & \yellowcell{41.19} & \yellowcell{50.81} & \bluecell{48.01} & \bluecell{53.84} \\
\bottomrule
\end{tabular}
}
\end{table*}

%% file: sections/tab_momentum.tex
\begin{table}[t]
\small
\centering
\caption{Study on exponential moving average (EMA).}
\label{tab:momentum}
\setlength{\tabcolsep}{3pt}
\begin{tabular}{ll cccc}
\toprule
\multicolumn{2}{l}{\textbf{Dataset}} & \textbf{Source} & \textbf{TENT} & \textbf{BATCLIP} & \textbf{MINT} \\
\midrule
\multirow{3}{*}{\rotatebox{90}{\scriptsize \textbf{ImageNet}}} 
& w/o EMA & 24.51 & 21.03 & 31.23 & 31.43 \\
& w/ EMA  & 25.87 & 24.98 & 31.90 & 32.15 \\
& $\Delta$     & \bluecell{+1.36} & \bluecell{+3.95} & \bluecell{+0.67} & \bluecell{+0.72} \\
\midrule
\multirow{3}{*}{\rotatebox{90}{\scriptsize \textbf{CIFAR10}}} 
& w/o EMA & 61.15 & 61.30 & 75.48 & 72.95 \\
& w/ EMA  & 68.87 & 68.18 & 77.79 & 76.33 \\
& $\Delta$     & \bluecell{+7.72} & \bluecell{+6.88} & \bluecell{+2.31} & \bluecell{+3.38} \\
\midrule
\multirow{3}{*}{\rotatebox{90}{\scriptsize \textbf{CIFAR100}}} 
& w/o EMA & 35.78 & 35.41 & 41.67 & 42.88 \\
& w/ EMA  & 40.12 & 41.13 & 43.87 & 44.49 \\
& $\Delta$     & \bluecell{+4.34} & \bluecell{+5.72} & \bluecell{+2.20} & \bluecell{+1.61} \\
\bottomrule
\end{tabular}
\end{table}

%% file: sections/tab_prompt.tex
\begin{table}[t]
\centering
\scriptsize
\caption{Sensitivity analysis with different prompt designs. \algname\ is robust to different prompt templates, consistently enhancing the baseline performance.}
\label{tab:prompt}
\setlength{\tabcolsep}{1.2pt}
\begin{tabular}{l ccc ccc ccc}
\toprule
\multirow{2}{*}{\textbf{Type}}
& \multicolumn{3}{c}{\textbf{ImageNet-C}}
& \multicolumn{3}{c}{\textbf{CIFAR10-C}}
& \multicolumn{3}{c}{\textbf{CIFAR100-C}} \\
\cmidrule(lr){2-4} \cmidrule(lr){5-7} \cmidrule(lr){8-10}
& Source & \algname & $\Delta$
& Source & \algname & $\Delta$
& Source & \algname & $\Delta$ \\
\midrule
T0 & 24.54 & 25.87 & +1.33
   & 61.16 & 68.87 & +7.71
   & 35.78 & 39.98 & +4.20 \\

T1 & 23.61 & 25.57 & +1.96
   & 62.05 & 63.86 & +1.81
   & 36.26 & 40.64 & +4.38 \\

T2 & 25.28 & 26.53 & +1.25
   & 61.19 & 68.80 & +7.61
   & 34.74 & 39.05 & +4.31 \\

T3 & 21.38 & 22.26 & +0.88
   & 60.14 & 70.09 & +9.95
   & 32.25 & 35.80 & +3.55 \\

T4 & 25.25 & 25.98 & +0.73
   & 60.43 & 68.05 & +7.62
   & 34.49 & 38.19 & +3.70 \\

T5 & 24.10 & 25.29 & +1.19
   & 64.21 & 65.41 & +1.20
   & 34.42 & 38.70 & +4.28 \\

T6 & 23.25 & 23.95 & +0.70
   & 61.33 & 68.94 & +7.61
   & 33.27 & 37.28 & +4.01 \\

T7 & 24.78 & 26.01 & +1.23
   & 61.45 & 69.03 & +7.58
   & 34.62 & 38.66 & +4.04 \\
\bottomrule
\end{tabular}
\end{table}

%% file: sections/tab_num_class.tex
\begin{table*}[htbp]
\centering
\caption{Sensitivity analysis to the number of classes. We sample subsets from the ImageNet-C dataset with the number of classes varying from 100 to 1,000.}
\label{tab:num_class}
\setlength{\tabcolsep}{4pt}
\begin{tabular}{ll cccccccccc}
\toprule
\multicolumn{2}{l}{\textbf{\# Class}} & \textbf{100} & \textbf{200} & \textbf{300} & \textbf{400} & \textbf{500} & \textbf{600} & \textbf{700} & \textbf{800} & \textbf{900} & \textbf{1000} \\
\midrule
\multirow{3}{*}{\textbf{Source}}
& Vanilla  & 25.35 & 25.52 & 25.66 & 25.20 & 25.29 & 25.13 & 24.88 & 24.55 & 24.38 & 24.54 \\
& \algname & 26.23 & 26.62 & 27.05 & 26.41 & 26.74 & 26.73 & 26.57 & 26.21 & 25.59 & 25.87 \\
& $\Delta$ & \bluecell{+0.88} & \bluecell{+1.10} & \bluecell{+1.39} & \bluecell{+1.21} & \bluecell{+1.45} & \bluecell{+1.60} & \bluecell{+1.69} & \bluecell{+1.66} & \bluecell{+1.21} & \bluecell{+1.33} \\
\midrule
\multirow{3}{*}{\textbf{TENT}}
& Vanilla  & 25.47 & 25.91 & 26.28 & 25.82 & 26.05 & 25.95 & 25.59 & 25.23 & 25.03 & 25.18 \\
& \algname & 26.30 & 26.77 & 27.13 & 26.60 & 26.99 & 26.64 & 26.04 & 25.53 & 25.06 & 24.98 \\
& $\Delta$ & \bluecell{+0.83} & \bluecell{+0.86} & \bluecell{+0.85} & \bluecell{+0.78} & \bluecell{+0.94} & \bluecell{+0.69} & \bluecell{+0.45} & \bluecell{+0.30} & \bluecell{+0.03} & \redcell{-0.20} \\
\midrule
\multirow{3}{*}{\textbf{MINT}}
& Vanilla  & 29.70 & 30.27 & 30.49 & 29.94 & 30.11 & 29.81 & 29.47 & 29.14 & 28.99 & 30.06 \\
& \algname & 28.64 & 31.07 & 32.33 & 32.47 & 32.81 & 32.58 & 32.51 & 32.27 & 31.99 & 32.15 \\
& $\Delta$ & \redcell{-1.06} & \bluecell{+0.80} & \bluecell{+1.84} & \bluecell{+2.53} & \bluecell{+2.70} & \bluecell{+2.77} & \bluecell{+3.04} & \bluecell{+3.13} & \bluecell{+3.00} & \bluecell{+2.09} \\
\midrule
\multirow{3}{*}{\textbf{BATCLIP}} 
& Vanilla  & 27.16 & 29.01 & 30.15 & 30.25 & 30.77 & 30.62 & 30.37 & 30.04 & 29.90 & 29.20 \\
& \algname & 28.64 & 30.97 & 32.26 & 32.38 & 32.76 & 32.48 & 32.41 & 32.13 & 31.78 & 31.90 \\
& $\Delta$ & \bluecell{+1.48} & \bluecell{+1.96} & \bluecell{+2.11} & \bluecell{+2.13} & \bluecell{+1.99} & \bluecell{+1.86} & \bluecell{+2.04} & \bluecell{+2.09} & \bluecell{+1.88} & \bluecell{+2.70} \\
\bottomrule
\end{tabular}
\end{table*}

%% file: sections/tab_latency.tex
\begin{table}[htbp]
\centering
\scriptsize
\caption{Running time (seconds) of two stages: SubTTA ($t_{\text{Sub}}$) and standard TTA ($t_{\text{TTA}}$).}
\label{tab:latency}
\setlength{\tabcolsep}{1.2pt}
\begin{tabular}{l ccc ccc ccc}
\toprule
\multirow{2}{*}{\textbf{Time}} & \multicolumn{3}{c}{\textbf{ImageNet-A}} & \multicolumn{3}{c}{\textbf{ImageNet-R}} & \multicolumn{3}{c}{\textbf{ImageNet-K}} \\
\cmidrule(lr){2-4} \cmidrule(lr){5-7} \cmidrule(lr){8-10}
 & TENT & BATCLIP & MINT & TENT & BATCLIP & MINT & TENT & BATCLIP & MINT \\
\midrule
$t_{\text{Sub}}$ & 0.16 & 0.17 & 0.15 & 0.16 & 0.16 & 0.15 & 0.15 & 0.16 & 0.15 \\
$t_{\text{TTA}}$ & 0.09 & 0.14 & 0.15 & 0.09 & 0.15 & 0.16 & 0.10 & 0.15 & 0.15 \\
\bottomrule
\end{tabular}
\end{table}

%% file: sections/checklist.tex
\section{Potential Risks}
The proposed \algname, while enhancing the robustness of VLMs under distribution shifts, relies on statistical properties of test data streams.
\algname\ estimates covariance matrices from incoming test batches to perform subspace alignment. 
As analyzed in our hyperparameter studies (Section~\ref{sec:exp-hyper}), extremely small batch sizes (e.g., $B=16$) may result in statistically unreliable covariance estimation, leading to performance fluctuations.
Additionally, like most TTA methods, our approach requires access to the model's visual and textual features during inference, which may limit applicability in strictly black-box API scenarios where feature embeddings are inaccessible.

\section{Use Or Create Scientific Artifacts}
Our work is built on established public benchmarks and pre-trained models. 
For evaluation, we use widely adopted corrupted image classification datasets including \texttt{CIFAR-10-C}, \texttt{CIFAR-100-C}, and \texttt{ImageNet-C}.
We do not modify the original content of these datasets. We utilize pre-trained CLIP models (\texttt{ViT-B-16}, \texttt{ViT-B-32}, \texttt{ViT-L-14}) as the backbone.
We will release the codebase of \algname\ upon publication to facilitate reproducibility and future research.

\subsection{Cite Creators Of Artifacts}
All external artifacts are properly credited to their original publications and repositories.

The benchmarks used in this work, including \texttt{CIFAR-10-C} and \texttt{CIFAR-100-C}~\cite{hendrycks2019benchmarking} (based on CIFAR~\cite{krizhevsky2009learning}), and \texttt{ImageNet-C}~\cite{hendrycks2019benchmarking} (based on ImageNet~\cite{deng2009imagenet}), are credited to their respective authors.

The CLIP~\cite{radford2021learning} backbones used in this work, including ViT-B-16. ViT-B-32, ViT-L-14, are referenced to its official publication.

\subsection{Discuss The License For Artifacts}
We comply with the licenses of all artifacts used in this work. The datasets (CIFAR-10-C, CIFAR-100-C, ImageNet-C) are available for research purposes under their respective licenses (typically Creative Commons or similar non-commercial research licenses). The CLIP models are released under the MIT License. Our own code will be released under a compatible open-source license (e.g., MIT).

\subsection{Artifact Use Consistent With Intended Use}
We confirm that our use of datasets and pre-trained models is consistent with their intended purpose. The corrupted datasets are specifically designed for benchmarking model robustness, which aligns directly with our goal of evaluating Test-Time Adaptation. The pre-trained CLIP models are used for zero-shot classification and feature extraction as intended by their creators.

\subsection{Data Contains Personally Identifying Info Or Offensive Content}
The datasets used (\texttt{CIFAR-10-C}, \texttt{CIFAR-100-C}, \texttt{ImageNet-C}) are standard computer vision benchmarks derived from public object classification datasets. To our knowledge, they do not contain sensitive personally identifying information or offensive content beyond what exists in the standard ImageNet/CIFAR distributions, which are widely accepted in the community.

\subsection{Documentation Of Artifacts}
We provide detailed documentation of all evaluation datasets and model combinations used in our experiments. Specifically, Appendix~\ref{sec:app-dataset} describes the datasets statistics, while Appendix~\ref{sec:app-exp} discusses the VLMs included along with their sizes and sources. Together, these materials ensure transparency regarding the domains, data characteristics, and model diversity involved in our study.

\section{Statistics For Data}
Dataset statistics are summarized as follows based on the details in Appendix~\ref{sec:app-dataset}.

\subsection{ImageNet-C}
\begin{itemize}[noitemsep, topsep=0pt]
  \item Number of classes: 1,000 object classes.
  \item Size: 50,000 test images for each corruption type and severity level.
  \item Structure: Derived from ImageNet validation set with 15 algorithmically generated corruptions.
\end{itemize}

\subsection{CIFAR-10-C}
\begin{itemize}[noitemsep, topsep=0pt]
  \item Number of classes: 10 distinct classes.
  \item Size: 10,000 test images.
  \item Structure: Derived from CIFAR-10 test set with 15 algorithmically generated corruptions.
\end{itemize}

\subsection{CIFAR-100-C}
\begin{itemize}[noitemsep, topsep=0pt]
  \item Number of classes: 100 fine-grained classes.
  \item Size: 10,000 test images.
  \item Structure: Derived from CIFAR-100 test set with 15 algorithmically generated corruptions.
\end{itemize}

\subsection{ImageNet-A}
\begin{itemize}[noitemsep, topsep=0pt]
    \item Number of classes: 200 distinct classes.
    \item Size: 7,500 test images.
    \item Structure: Composed of naturally occurring adversarial examples (real-world, unmodified images) that standard ResNet models consistently misclassify, designed to evaluate model robustness.
\end{itemize}

\subsection{ImageNet-R}
\begin{itemize}[noitemsep, topsep=0pt]
    \item Number of classes: 200 distinct classes.
    \item Size: 30,000 test images.
    \item Structure: Contains various artistic renditions of ImageNet objects, including art, cartoons, paintings, origami, toys, and video game graphics, to test out-of-distribution generalization to different textures and styles.
\end{itemize}

\subsection{ImageNet-K}
\begin{itemize}[noitemsep, topsep=0pt]
    \item Number of classes: 1,000 object classes.
    \item Size: 50,000 test images.
    \item Structure: Consists of black-and-white sketch images corresponding to the original ImageNet classes, collected via web search to evaluate model robustness to severe domain shifts.
\end{itemize}

\section{Computational Experiments}\label{sec:app-exp}
All computational experiments in this work are fully reproducible, with details provided in Section~\ref{sec:exp-setup}.

\subsection{Model Size And Budget}
We evaluate our method on CLIP models with varying capacities, including:
\begin{itemize}[noitemsep, topsep=0pt]
    \item \texttt{ViT-B-16}: 112M total parameters, and 41k trainable parameters.
      \item \texttt{ViT-B-32}: 113M total parameters, and 41k trainable parameters.
      \item \texttt{ViT-L-14}: 343M total parameters, and 104k trainable parameters.
\end{itemize}
All experiments are executed on NVIDIA A100 80GB GPUs.

\subsection{Experimental Setup And Hyper-params}
We describe experimental settings in Section~\ref{sec:exp-setup}. We adopt an online adaptation setting where the model adapts to a continuous stream of unlabeled test data for each corruption type independently (resetting between corruptions).
Key hyperparameters studied in Section~\ref{sec:exp-hyper} include:
\begin{itemize}[noitemsep, topsep=0pt]
    \item Subspace rank $r$: We vary $r$ (e.g., 64, 128, 256) and find that avoiding extremes (too compressed or full-rank) is optimal.
    \item Momentum coefficient $\alpha$: We use an EMA strategy for covariance updates, with stability observed for $\alpha \in (0.2, 0.8)$.
    \item Batch Size: We investigate the impact of batch size, and results indicate that larger batch sizes (e.g., 64 and above) provide stable covariance estimation and consistent performance, whereas smaller batches (e.g., 16) may lead to instability.
\end{itemize}

\subsection{Descriptive Statistics}
We report the classification accuracy as the evaluation metric. We report the mean performance across 15 corruption types at severity level 5.

\subsection{Parameters For Packages}
The existing packages used are specified as follows. We use \texttt{PyTorch} (v2.2.1) as the core deep learning framework, together with \texttt{open-clip-torch} (v2.24.0) and \texttt{huggingface-hub} (v0.21.4) for model implementation and text pre-processing.

\section{AI Assistants In Research Or Writing}
In this paper, AI assistant tool is used to edit and improve the quality of the text, including checking the spelling, grammar, punctuation and clarity.